\documentclass[runningheads]{llncs}

\usepackage[T1]{fontenc}    
\usepackage[utf8]{inputenc}     
\usepackage[english]{babel}
\usepackage{multirow}
\usepackage[justification=centering]{subfig}
\usepackage{graphicx}
\usepackage{amsmath}
\usepackage{amssymb}
\usepackage{mathtools} % for command \coloneqq
\usepackage{booktabs}
\usepackage{svg}

\newcommand{\R}{\mathbb{R}}

\newcommand{\pred}{\mathrm{pred}}
\newcommand{\gt}{\mathrm{gt}}
\DeclareMathOperator*{\argmax}{argmax}
\DeclareMathOperator{\divergence}{div}
\newcommand{\enstq}[2]{\left\{ #1 \mathrel{} \middle\vert \mathrel{} #2 \right\}}

\makeatletter
\DeclareRobustCommand\onedot{\futurelet\@let@token\@onedot}
\def\@onedot{\ifx\@let@token.\else.\null\fi\xspace}

\def\eg{\emph{e.g}\onedot} 
\def\ie{\emph{i.e}\onedot} 
 
\def\etc{\emph{etc}\onedot}

\def\etal{\emph{et al}\onedot}
\makeatother

\usepackage[ruled,vlined,lined,linesnumbered]{algorithm2e}
\SetKwInOut{Input}{input}
\SetKwInOut{Parameter}{parameter}
\SetKwInOut{Output}{output}
\SetKwComment{Comment}{}{}

\SetCommentSty{mycmtsty}

\usepackage{hyperref}
\usepackage{color}

\usepackage[capitalize]{cleveref} % Needs to be after hyperref
\crefname{section}{Sec.}{Secs.}
\Crefname{section}{Section}{Sections}
\crefname{appendix}{App.}{Apps.}
\Crefname{appendix}{Appendix}{Appendices}
\crefname{table}{Tab.}{Tabs.}
\Crefname{table}{Table}{Tables}
\crefname{algorithm}{Alg.}{Algs.}
\Crefname{algorithm}{Algorithm}{Algorithms}

\usepackage{etoolbox}
\usepackage{refcount} % this allows \getrefnumber{} instead of \ref{}, which does not put a hyperlink
\newcommand{\repthanks}[1]{\textsuperscript{\getrefnumber{#1}}}
\makeatletter
\patchcmd{\maketitle}
  {\def\thanks}
  {\let\repthanks\repthanksunskip\def\thanks}
  {}{}
\patchcmd{\@maketitle}
  {\def\thanks}
  {\let\repthanks\@gobble\def\thanks}
  {}{}
\newcommand\repthanksunskip[1]{\unskip{}}
\makeatother

\begin{document}

\title{Improving OCR using internal document redundancy}

\author{Diego Belzarena\inst{1,2}\thanks{D.\ Belzarena and S.\ Mowlavi contributed equally to this work.\protect\label{equalcontrib}} \and
Seginus Mowlavi\inst{1}\repthanks{equalcontrib} \and
Aitor Artola\inst{3} \and
Camilo Mariño\inst{1,2} \and
Marina Gardella\inst{1} \and
Ignacio Ramírez\inst{2} \and
Antoine Tadros\inst{4,1} \and
Roy He\inst{3} \and
Natalia Bottaioli\inst{1} \and
Boshra Rajaei \inst{5} \and
Gregory Randall\inst{2} \and
Jean-Michel Morel\inst{3}
}
\authorrunning{D. Belzarena, S. Mowlavi et al.}
\institute{Université Paris-Saclay, ENS Paris-Saclay, Centre Borelli, Gif-sur-Yvette, France \and
IIE, Facultad de Ingenería, Universidad de la República,
Uruguay \and
City University of Hong Kong, Kowloon, Hong Kong \and
Determinant France, Paris, France \and
Sadjad University, Mashhad, Iran
}

{
\renewcommand{\thefootnote}{\fnsymbol{footnote}}
\maketitle
}

\begin{abstract}
Current OCR systems are based on deep learning models trained on large amounts of data. Although they have shown some ability to generalize to unseen data, especially in detection tasks, they can struggle with recognizing low-quality data. This is particularly evident for printed documents, where intra-domain data variability is typically low, but inter-domain data variability is high. In that context, current OCR methods do not fully exploit each document's redundancy. We propose an unsupervised method by leveraging the redundancy of character shapes within a document to correct imperfect outputs of a given OCR system and suggest better clustering. To this aim, we introduce an extended Gaussian Mixture Model (GMM) by alternating an Expectation-Maximization (EM) algorithm with an intra-cluster realignment process and normality statistical testing. We demonstrate improvements in documents with various levels of degradation, including recovered Uruguayan military archives and 17th to mid-20th century European newspapers.
\end{abstract}

\section{Introduction} \label{sec:intro}

Optical Character Recognition (OCR) is an important problem in computer vision, and it has practical applications in various fields, such as the humanities and computational linguistics. This challenging task has been key to promoting far-reaching discoveries~\cite{lecun1989backpropagation}.

A major technical hurdle hampering OCR performance is that the current deep learning technology does not detect errors. We argue that these errors can be avoided or corrected by exploiting the currently overlooked internal document redundancy. Redundancy refers to consistently using the same languages and fonts, and the image quality within similar documents.

At the heart of its difficulty lies the fact that OCR systems typically need to run at the word or line granularity level, using Hidden Markov Model~\cite{el1999hmm}, Connectionist Temporal Classification~\cite{graves2008ctc}, or Transformer~\cite{vaswani2017attention} decoders. These systems output text strings without precisely localizing individual characters, bypassing the character segmentation issue, which has proven to be very difficult since the days of rule-based OCR algorithms~\cite{smith2007tesseract}. Successful applications of Convolutional Neural Networks (CNN) to image classification~\cite{krizhevsky2012imagenet} and segmentation~\cite{ronneberger2015unet} have traded their unprecedented power with a new barrier to character segmentation: they rely on annotated data, which are prohibitively costly to produce at the character level. Meanwhile, abundant research has integrated CNNs into word or line-level methods to tackle the challenging handwriting settings~\cite{bluche2017htr} and scene texts~\cite{long2021scene}. Since then, deep learning has remained at the forefront of OCR technology, with the Transformer architecture unlocking coarser granularity (paragraph, entire image) for greater versatility~\cite{coquenet2023dan}.

Consequently, the philosophy on generalizability has moved toward acquiring and increasing training data~\cite{volpi2018generalizing}: shapes in new input images are recognized by their similarity to learned patterns, not by comparing them to patterns emerging within the input data. Notably, the generalization power is hardly interpretable, and even state-of-the-art models can fail without warning in unforeseen situations, hindering trust in the outputs. This approach remains as advances in OCR trickle down to the classical setting of printed text. Here, we make the case for exploiting the inherent redundancy among character shapes. Our contributions can be summarized as follows:
\begin{itemize}
    \item We propose improving OCR results by normalizing cropped character images and clustering them according to an extended Gaussian Mixture Model (GMM), where iterative sample image registration steps and cluster normality tests refine clusters. This method, built on general principles, learns the shapes of characters in a document without supervision, yielding easily interpretable outputs.
    \item We implement a proof-of-concept and conduct experiments on historical printed and typewritten documents. The code is publicly available at \url{https://github.com/seginusmowlavi/ocr-using-shape-redundancy}. Our results show that existing approaches under-use the internal redundancy of documents.
    \item Some of our experiments are performed on a new annotated dataset of typewritten documents from real archives in a wide range of preservation quality. The dataset is publicly available at \url{https://github.com/camilomarino/ocr_berrutti_dataset}.
\end{itemize}

Throughout the paper, we frequently use the terms ``symbol'', ``glyph'', and ``character'', whose general meaning can be ambiguous. In our usage, a \emph{symbol} designates a basic unit of text (letter, punctuation sign, \etc, except space); a \emph{glyph} is an abstract graphic representation of a symbol (different typefaces and fonts produce different glyphs of the same symbol). A \emph{character} is the realization of a glyph as a digital image. Therefore, character recognition is viewed as an image classification problem: characters are the input data, and symbols are the output labels. We make one exception for the Character Error Rate (CER) metric, which counts symbol errors according to our terminology but is an already established term.

\section{Related work} \label{sec:relatedwork}

\paragraph{Document redundancy}
 Document redundancy, namely the repetition of characters in an almost identical printed form, has been considered in the computer vision era preceding deep learning. For example, Kopec and Lomelin~\cite{kopec1997supervised} use a two-phase iterative training algorithm alternating between a transcription alignment and an aligned template estimation step at each iteration. The latter step involves assigning template pixel colors to maximize the likelihood. This enables the visualization of transcription errors and printing font generation. Both goals remain valid today. More recently, Siglidis \etal~\cite{siglidis2024learnable} have proposed learning to reconstruct images to discover the underlying glyphs, but their method does not provide a stochastic analysis of character variability.

\paragraph{Character-level OCR}

To bypass the cost of annotating data at the character granularity level, CharNet~\cite{xing2019charnet} and CRAFT~\cite{baek2019craft} devise an iterative weakly supervised training scheme where the network-in-training itself performs annotations. To our knowledge, this approach has not been further explored. Other works sidestep the problem with generative approaches by learning a model of characters' image formation, with~\cite{berg2013ocular} or without~\cite{siglidis2024learnable} guidance from a language model.

\paragraph{Post-OCR correction and globality}

Traditional OCR systems make limited or no use of available prior information from the language they are transcribing. Therefore, a large body of work has been dedicated to the so called ``post-OCR correction'' task~\cite{nguyen2021survey,rigaud2019icdar},  where language models are used to fix OCR outputs. The recent development of attention mechanisms in OCR architectures has implicitly or explicitly integrated language modeling in an end-to-end fashion, enabling the OCR to process inputs globally at the scale of the document~\cite{coquenet2023dan,li2023trocr}. Even though attention has been demonstrated to use global textual information within a document~\cite{bottaioli2024normalized,constum2024end}, whether it leverages the document's visual redundancy is yet to be substantiated. The experiments presented in this paper suggest that visual redundancy is under-used. A post-processing approach that exploits this visual information would be suitable to address this issue and would offer complementarity with the existing NLP-based post-OCR correction.

\section{Correcting OCR with character image clustering} \label{sec:clustering}

 We propose a simple method to improve the output of any OCR algorithm (which we call the \emph{base OCR}) by clustering character images. The clusters' empirical mean and covariance provide a model for each glyph. Accurate clustering enables error correction by the rule of majority, based on the assumption that the base OCR has a higher-than-random recognition performance when restricted to any subset of the alphabet. Note that the end goal is correcting labeling errors and not clustering for its sake, so we can, for example, allow our clusters to over-segment the set of images (\ie partition the set into more clusters than needed).

Our method is meant to be applied to any large enough series of images of printed or typewritten documents from a typographically consistent source, such as books, newspapers, or archives. It is based on the hypothesis that characters of the same symbol, typeface, and font (\ie occurrences of the same glyph) follow a Gaussian distribution. We make this assumption in the image space, consequently our method does not require transforming the images into any feature space, easing the visual interpretability of the results.

However, this Gaussianity assumption relies strongly on the correct alignment of the samples between themselves. In this setting, image alignment poses somewhat of a chicken-and-egg problem: alignment can be performed by registering similar images, but this assumes that they have already been correctly clustered. Conversely, one cannot hope to cluster the image space without some previous image alignment. We tackle this challenge by proceeding in three steps. First, all character images of fixed dimension are extracted using the bounding boxes predicted by the OCR, and roughly aligned (registered). Then, they are clustered following a Gaussian Mixture Model. Finally, each cluster is iteratively registered, tested and divided into sub-clusters if it fails a Gaussianity test.

The following \Cref{ssec:preprocessing,ssec:GMM,ssec:trees} provide more details about the above steps, while \Cref{ssec:relabeling} explains the process of correcting character recognition errors from the clustering result.

\subsection{Character bounding-box pre-processing and image standardization} \label{ssec:preprocessing}

The first step is to convert the base OCR's character bounding boxes into standardized images. The naive way of cropping and resizing to fixed dimensions $H\times W$ would make each character's perceived size dependent on the tightness of the bounding box. Instead, we pad each crop to match the dimensions $H\times W$, only down-scaling bigger crops by a discrete factor (a power of a fixed parameter $s>1$) as needed. We choose the size of padding margins so that the resulting image, seen as a distribution of mass (with mass at blacks, no mass at whites) have their barycenter at coordinates $(W/2,H/2)$.

Additionally, we perform radiometric and geometric pre-processing before cropping. Radiometric pre-processing tackles the issue of background discontinuity when padding, as the crops' background can be non-uniformly gray and noisy. We remove the background by threshold masking, and use Poisson editing \cite{perez2023poisson} to make the text blend seamlessly with the new white background. Geometric pre-processing aims at correcting the bounding boxes, as they can be too tight (cutting away pieces of characters) or too loose (including fragments of neighboring characters). We use a procedure involving minimal cost paths \cite{dijkstra2022note} to find the outline of characters, outputting precise masks to be used in conjunction with cropping. Both pre-processing algorithms, which do not involve learning and are easily interpretable, are detailed in the supplementary material.

\begin{figure}[ht]
    \centering
    \includegraphics[width=0.6\linewidth]{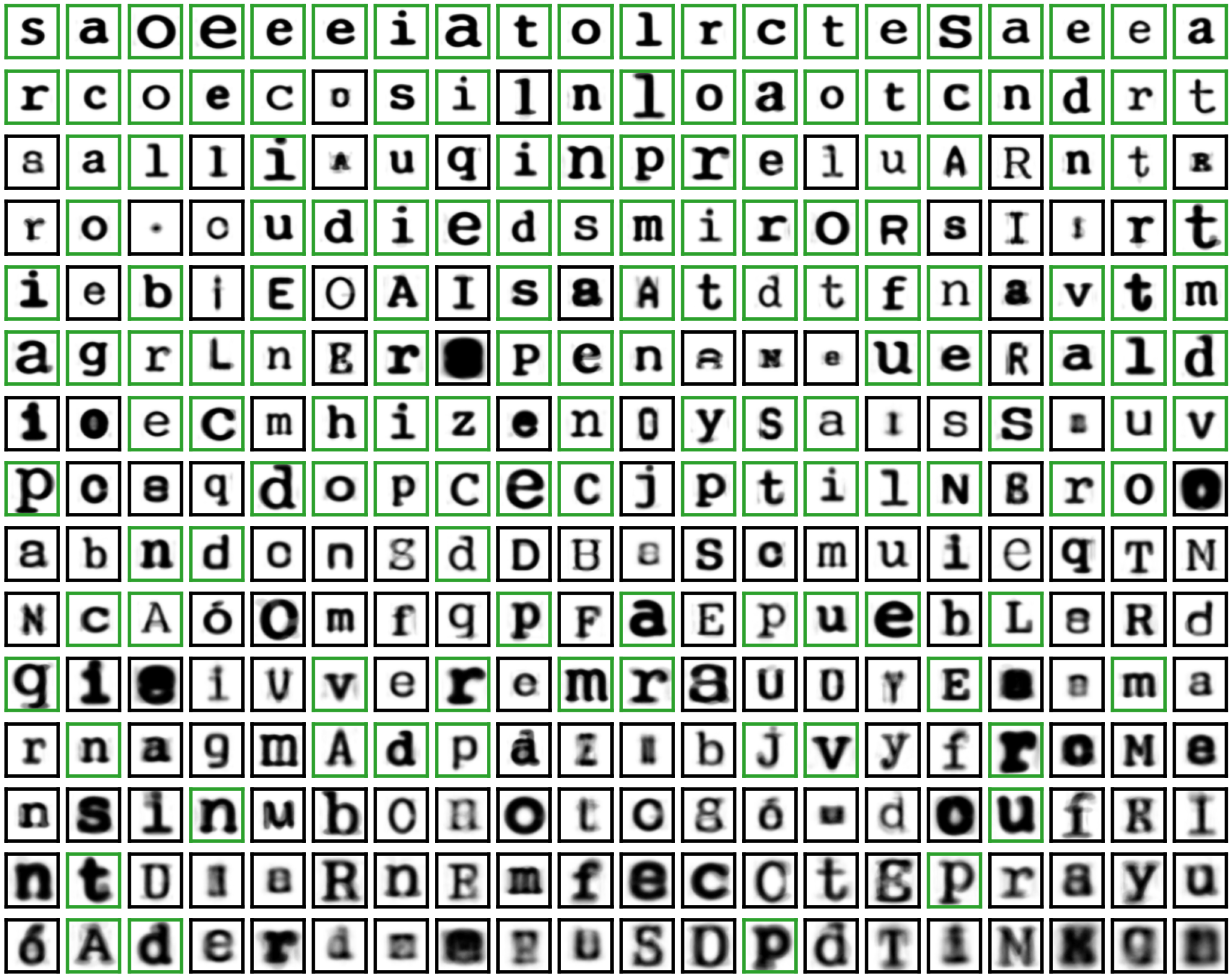}
    \caption{Cluster means after applying a GMM to all preprocessed character images of the Berrutti OCR collection extracted by CharNet, here with $K=300$ clusters. Clusters where more than 90\% of samples have the same predicted label are painted green. The clusters are sorted in increasing order of total variance (the trace of the covariance matrix). We observe that some cluster means blend two or more glyphs, especially in the mosaic's lower portion.}
    \label{fig:global_GMM}
\end{figure}

\subsection{GMM clustering} \label{ssec:GMM}

To speed up computation, images are projected from $\R^{H\times W}$ to a $D$-dimensional space by principal component analysis (PCA), where $D$ is chosen so that the selected dimensions retain a given fraction $q_\mathrm{variance}$ of the total variance, \ie the sum of the eigenvalues of the global covariance matrix along each dimension.

The GMM model in $\R^D$ is initialized with a preset number $K$ of components using the $k$-means algorithm, and further adjusted using the Expectation-Maximization (EM) algorithm on the projected images. Covariance matrices are estimated using Oracle Approximating Shrinkage (OAS)~\cite{chen2010shrinkage} to better handle clusters with few samples, which is important given that symbol frequencies are expected to follow Zipf's law \cite{piantadosi2014zipf}. After each M step of the EM algorithm, components with at most one sample are ignored, so the resulting number of components in the GMM is $C\leq K$. The parameters of the components are not saved for later stages of the method, as only the clustering matters here.

\subsection{Cluster refinement as a binary tree} \label{ssec:trees}

\Cref{fig:global_GMM} shows the means of the GMM components after applying the EM algorithm. Closely examining these means indicates that some of the corresponding elements are multi-modal, blending in more than one glyph. We propose an algorithm to split such multi-modal components into uni-modal ones so that all the components in the final model are uni-modal and fit the Gaussian assumption; thus, samples can be unambiguously registered to their means.

This algorithm (for which we provide a pseudo-code implementation in the supplementary material), starts from each component of the GMM and builds a binary tree of sub-clusters. At each node, images are registered to the cluster mean using a single-scale Inverse Compositional Algorithm (ICA)~\cite{baker2001equivalence,briand2018improvements}. The resulting set of registered images is then tested for normality. We experimentally observed that the multi-modal nature of composite clusters tends to appear in the first principal components. Accordingly, a fixed number $k$ of principal components are kept and are given a $p$-value from their Anderson-Darling normality statistic~\cite[\S4.8.1]{dagostino2017tests}. If any of those does not pass a threshold $p_\mathrm{thr}$, the node is split into two children by fitting a two-component GMM.

Trees are grown from each cluster until all leaves pass the Gaussianity test. To ensure the significance of the statistical testing, leaves that have fewer than $n_\mathrm{min}$ elements are discarded; the others are taken as the final clusters. Note that those clusters, therefore, do not strictly make up a partition of the set of samples; however, as experimentally shown in \Cref{sec:experiments}, they concern the vast majority of samples, including a significant number incorrectly recognized by the base OCR. 

\begin{figure}[h]
    \centering
    \subfloat[]{\label{subfig:tree1}\includegraphics[height=1cm]{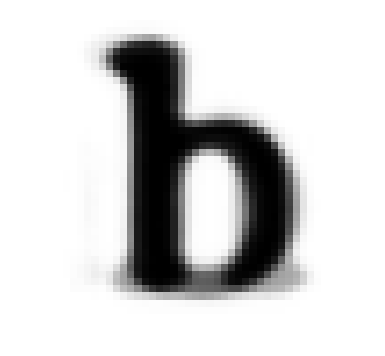}}
    \quad
    \subfloat[]{\label{subfig:tree2}\includegraphics[height=1cm]{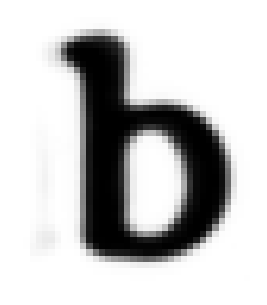}\includegraphics[height=1cm]{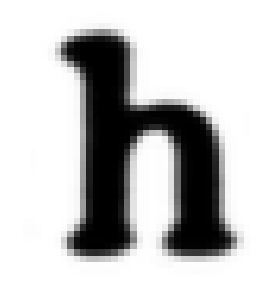}}
    \quad
    \subfloat[]{\label{subfig:tree3}\includegraphics[height=1cm]{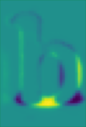}\,\includegraphics[height=1cm]{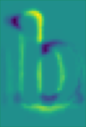}\,\includegraphics[height=1cm]{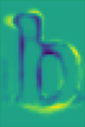}}
    \quad
    \subfloat[]{\label{subfig:tree4}\includegraphics[height=1cm]{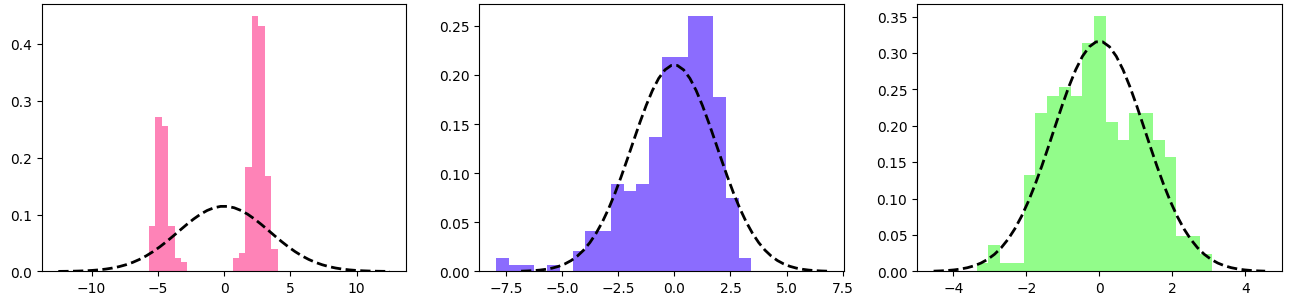}}
    \caption{Cluster with 64\% and 36\% of `b' and `h' images, respectively, through the refinement process. \protect\subref{subfig:tree1} Mean of cluster images. \protect\subref{subfig:tree2} Mean of sub-clusters images after refinement; the refinement perfectly separates the `b' and the `h'. \protect\subref{subfig:tree3} First three eigenvectors of the initial (before refinement) cluster's covariance matrix. \protect\subref{subfig:tree4} First three PCA projections; Anderson-Darling testing gives them $p$-values of $3.0\times 10^{-88}$, $6.3\times 10^{-12}$ and $2.9\times 10^{-2}$ respectively.}
    \label{fig:tree}
\end{figure}

\Cref{fig:tree} summarizes this procedure on a bimodal cluster produced by the global GMM of step \ref{ssec:GMM} (with modes `b' and `h'), resulting in a simple tree with two leaves. Note the proximity of the two modes when considered among all the GMM-processed images, which the global PCA can exacerbate. Performing a PCA solely on this cluster shows how the first principal component discriminates `b' images from the `h' ones (\cref{subfig:tree4}). This can be observed in the first eigenvector of the cluster covariance matrix (corresponding to the first principal component), which appears to be the difference between the ``h'' and ``b'' means (\cref{subfig:tree3}). 

\subsection{Correcting OCR errors} \label{ssec:relabeling}

The previous steps produced clusters of characters representing the same glyph. We use this information to correct the base OCR's recognition outputs according to the "super-majority wins over minority" principle. Specifically, for each cluster, we consider the predictions of the OCR (a symbol for each character) and their apparition frequency. We then replace each symbol by the one with the highest frequency (minority replaced by relative majority), provided that this frequency is higher than a preset threshold (super-majority needed) $f_\mathrm{thr}$; otherwise, we do not override the OCR's prediction for the cluster. A pseudo-code implementation is provided in the supplementary material. 

The following reasoning motivates this procedure. Assuming the base OCR is consistently better than a random guess, a uni-modal cluster would theoretically be correctly labeled by a majority vote. In practice, the clustering is imperfect, and a few clusters remain aggregations of images of different glyphs. Fortunately, the existing OCR systems are significantly better than random guesses, allowing us to filter out presumably bad clusters by requiring a super-majority.

\section{Experimental setup} \label{sec:experimentalSetup}

\subsection{Overview}

Here we present evidence that glyph image models can be learned from a large enough collection of documents from the same printing process. We show that this can improve the results of OCR systems, with the additional advantage of explaining each recognition. We shall, therefore, perform our experiments on datasets that satisfy the following criteria:
\begin{itemize}
    \item Domain-wise, the dataset should consist of images of printed (or typewritten) text, which are partitioned into sub-collections of pictures from the same or similar printing sources.
    \item Granularity-wise, the data and their annotation should allow for a straightforward evaluation process. Ideally, annotations should be at the character level to use a classification or object detection metric. To our knowledge, no dataset meets this requirement. Instead, we work with datasets where images show a single line of text, with their annotation being the ground-truth transcription; then, the sequential layout of the text in such images allows for the character-level output to be unambiguously assembled as a string which can then be evaluated using the standard CER metric.
\end{itemize}
The first criterion provides the setting where we claim an improvement, and the second makes evaluating our claim possible. We use two datasets meeting these criteria (see details in \cref{ssec:datasets}). A unified evaluation protocol is laid out in \Cref{ssec:evaluation}.

\subsection{Datasets} \label{ssec:datasets}

\paragraph{Berrutti OCR collection}

We use lines extracted from a subset of documents of the Berrutti Archive, a set of scans of typewritten documents from the dictatorship that ruled Uruguay between 1973 and 1985 \cite{hudson1992uruguay}. The original documents were microfilmed and maintained for decades in boxes until the Minister of Defense Azucena Berrutti discovered them in 2006 (hence the name). The microfilm rolls were scanned, producing more than 2 million images. A group of 175 images that comply with the Uruguayan law on protecting personal data was selected to represent the type and quality of the documents in the Berrutti Archive. The documents were preprocessed with noise reduction and alignment steps. At least one human expert manually corrected the Tesseract OCR output to produce ground truth transcriptions. Then, a human expert manually annotated the lines in each document using the VGG Image Annotator tool~\cite{dutta2019vgg}. The bounding boxes of these lines were then used to produce cropped images of each text line. Each document was manually classified into one of three categories--high, medium, and low quality-- according to a subjective appreciation of how challenging the automatic transcription of the document is. This subjective classification was consistent with the performance of several OCR methods when applied to those collections (see the detailed results in the supplementary material). The three resulting annotated sub-collections are described in \Cref{subtab:Berrutti_DDBB}.

\begin{table}[!htbp]
    \centering
    \subfloat[]{\label{subtab:Berrutti_DDBB}
        \resizebox{0.4\columnwidth}{!}{\begin{tabular}{cccc}
        \toprule
        Sub-collection   & NbP  & NbL & NbC \\
        \midrule
        Low Quality  & 46 & 3223 & 101048 \\
        Medium Quality  & 55 & 2352 & 107633 \\
        High Quality  & 74 & 2987 & 135106 \\
        \bottomrule
        \end{tabular}}
    }
    \subfloat[]{\label{subtab:ENP_DDBB}
        \resizebox{0.55\columnwidth}{!}{\begin{tabular}{lccc}
        \toprule
        Sub-collection  & Language & NbL & NbC \\
        \midrule
        1882/05/03 Alg.\ Han. & Dutch& 697 & 24778 \\
        1845/09/13 The Exa. &  English & 2390 & 100084 \\
        1940/09/09 Postimees &  Estonian &  2426 & 80645 \\
        1906/12/06 Tur.\ San.  & Finnish  & 520 & 14546 \\
        1894/12/23 Le Pet.\ Par.  & French  & 2423 & 78352 \\
        1928/08/04 Ham.\ Anz. & German  & 789 & 16896 \\
        1931/08/27 Jau.\ Vor. & Latvian &  583 & 18375 \\
        1939/09/01 Exp.\ Por.  & Polish  & 2645 & 69671 \\
        1910/04/17 Åbo Und. & Swedish  & 1671 & 46137 \\
        \bottomrule
        \end{tabular}}
    }
    \caption{\protect\subref{subtab:Berrutti_DDBB} Berrutti OCR collection public dataset. \protect\subref{subtab:ENP_DDBB} Used sub-collections from the ENP dataset with different languages, identified by issue date and abbreviated title. NbP: number of pages, NbL: number of lines, NbC: number of characters.}
    \label{tab:DDBB}
\end{table}

\begin{figure}[t]
    \centering
    \includegraphics[width=0.9\columnwidth]{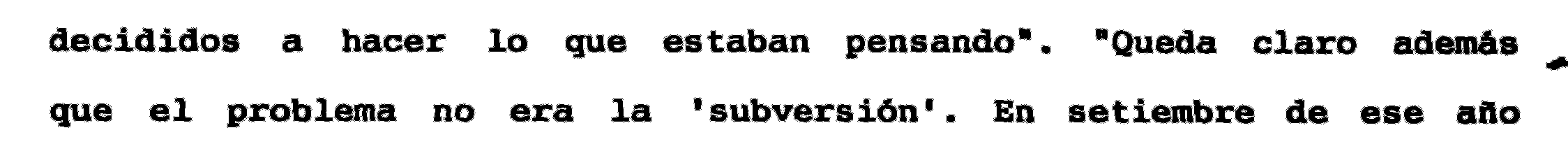}
    \includegraphics[width=0.9\columnwidth]{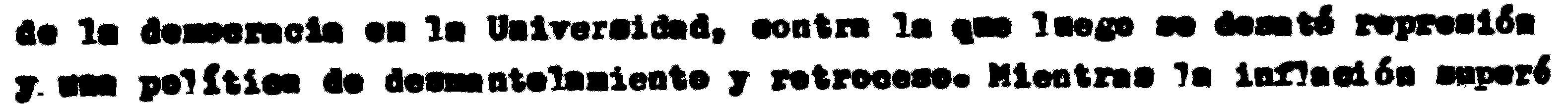}
    \includegraphics[width=0.9\columnwidth]{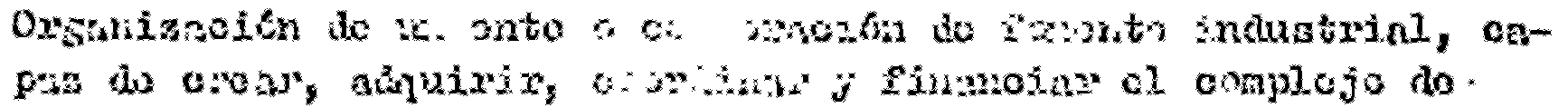}
    \includegraphics[width=0.9\columnwidth]{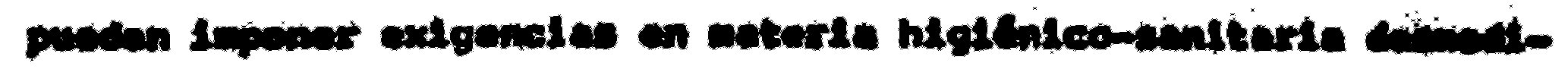}
    \caption{Examples of the Berrutti OCR collection images, with typical typewritten artifacts, such as irregular strength due to varying amounts of ink or stroke force and interaction between paper grain and ink diffusion. Other common artifacts are the presence of seals and dirt.}
    \label{fig:ejemplosBerrutti} 
\end{figure}

The documents in this collection were mainly produced by typewriter machines, which were common at the time, and the most typical problems concern the excess or lack of ink in the tape. \Cref{fig:ejemplosBerrutti} shows some examples. This publicly available dataset, in Spanish, is similar to the documents that populated archives in other parts of the continent at that time. It can be found at \url{https://github.com/camilomarino/ocr_berrutti_dataset}.

\paragraph{Europeana Newspapers Project}

The Europeana Newspapers Project (ENP) is a dataset of scans of European newspapers from the 17th to the mid-20th centuries. The data is human-annotated at the paragraph level and includes the line-level output of a commercial OCR engine. We use this information to build a set of annotated cropped images of lines for each document; to mitigate the OCR engine errors, we automatically filter out lines that do not meet an annotation quality criterion.

\begin{figure}[!htbp]
    \centering
    \includegraphics[width=0.9\columnwidth]{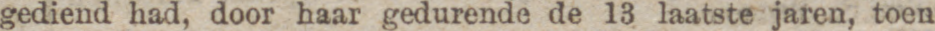}
    \includegraphics[width=0.9\columnwidth]{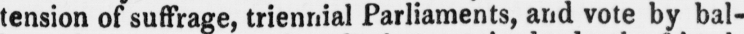}
    \includegraphics[width=0.9\columnwidth]{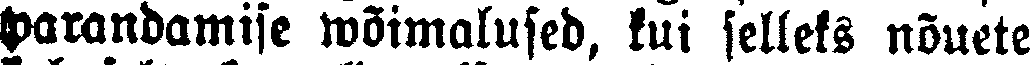}
    \includegraphics[width=0.9\columnwidth]{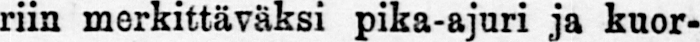}
    \includegraphics[width=0.9\columnwidth]{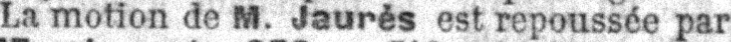}
    \includegraphics[width=0.9\columnwidth]{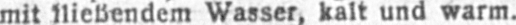}
    \includegraphics[width=0.9\columnwidth]{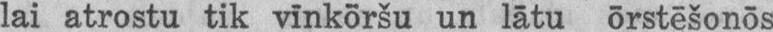}
    \includegraphics[width=0.9\columnwidth]{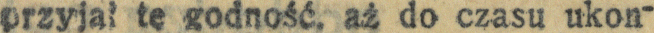}
    \includegraphics[width=0.9\columnwidth]{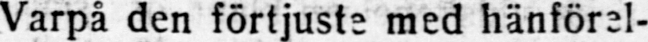}
    \caption{Examples of the ENP collection.}
    \label{fig:ejemplosENP} 
\end{figure}

For our experiments, we keep a subset of documents from different countries and with a diversity of languages, as described in \Cref{subtab:ENP_DDBB}, grouped in sub-collections by newspaper issue (title and date). The size of the sub-collections permits the definition of image models for each glyph. As seen in \Cref{fig:ejemplosENP}, these newspapers use a variety of fonts. Unlike the Berrutti OCR collection, which is formed by mono-space documents produced by typewriter machines, the ENP collection is of printed material. Hence, its documents have varying spaces between words, a fundamental difference between both collections.
        
\subsection{Evaluation protocol and metrics}
\label{ssec:evaluation}

For evaluation on a single input line image, each method is asked to output a set of detections $D_\pred$, where detection is understood to be a bounding box with its predicted symbol. This is needed to ensure a fair comparison with our proposed post-processing, which acts on the output by correcting $D_\pred$.

To score $D_\pred$ against the ground-truth string $s_\gt$ using the CER metric, we must reconstruct a string $s_\pred$ from $D_\pred$. This means ordering the elements of $D_\pred$. We proceed by sorting the centers of the bounding boxes from left to right along the $x$-axis. Note that $s_\pred$ does not include spaces, as we cannot predict their positions trivially from $D_\pred$. Thus, spaces are removed from $s_\gt$.

Annotations sometimes take advantage of the wealth offered by the Unicode standard. To remove as many unwanted ambiguities as possible, we further standardize $s_\mathrm{pred}$ and $s_\mathrm{gt}$ by applying Unicode's Normalization Form C~\footnote{https://unicode.org/reports/tr15/} and replacing symbols with semantically differing variants (hyphens and quotation marks) with their ASCII counterpart.

Once all these operations are performed, the CER metric is applied:
\begin{equation} \label{eq:cer}
	\mathrm{CER}(s_\mathrm{gt}, s_\mathrm{pred}) = \frac{I+D+S}{\mathrm{length}(s_\mathrm{gt})}
\end{equation}
where $I,D,S\in\mathbb{Z}_{\geq0}$ are such that one can transform $s_\mathrm{gt}$ into $s_\mathrm{pred}$ using $I$ insertions, $D$ deletions and $S$ substitutions of symbols, for which \eqref{eq:cer} is minimal.

The resulting protocol differs slightly from what is commonly used in text recognition works because spaces are removed from transcriptions before comparison. This should not be viewed as an issue since the purpose is not to assess absolute OCR quality but to use the CER metric as a proxy to evaluate the performance of our proposed post-processing method on a character basis.

\subsection{Implementation details}

We apply our method on two different baseline OCR systems: CharNet \cite{xing2019charnet} (specifically designed at the character level) and Google Cloud Vision's Document OCR \cite{gcv} (a state-of-the-art commercial engine). The used parameters are: $H=48$, $W=32$, $s=1.2$ (\cref{ssec:preprocessing}); $q_\mathrm{variance}=0.9$, $K=700$ (\cref{ssec:GMM}); $n_\mathrm{min}=20$, $p_\mathrm{thr}=\mathbb{P}_{X\sim\mathcal{N}(0,1)}(\lvert X\rvert>2)\approx0.0455$, $k=9$ (\cref{ssec:trees}); and $f_\mathrm{thr}=0.6$ (\cref{ssec:relabeling}).

\section{Experiments} \label{sec:experiments}

We present the outcome of experiments based on the setup detailed in \Cref{sec:experimentalSetup}. While we base our observations on the outputs of the character relabeling step (\cref{ssec:relabeling}), 
we do not aim to push it as a definitive answer to OCR post-processing. Instead, these results should be taken as evidence of the approach's ability to learn the image models of the glyphs from the image and its explainability, interpretability, and complementarity with existing OCR systems.

\subsection{Quantitative results} \label{ssec:quant}

\Cref{subtab:cer} presents the quantitative results of two reference character-level OCR methods and the effect of the proposed post-processing method. It shows an improvement from the CharNet baseline, shaving more than one percentage point of CER on either Berrutti or ENP. Note that our method is only designed to correct character \emph{substitution} errors, not \emph{insertions} nor \emph{deletions}. Estimating these respectively using $S$, $I$, and $D$ of \Cref{eq:cer} shows that CharNet recognition errors account for around $7.5$ (resp. $10$) percentage points of its error rate on Berrutti (resp. ENP). Hence, one can reasonably conclude that our method achieves more than $1/10$th of the maximum theoretical gain. On the other hand, results from the state-of-the-art Google baseline, which has a much better performance than CharNet, show non-negligible increases in CER. For this case, our method introduces more errors than corrections, giving it limited applicability as a fully automated OCR post-processor. This does not undermine the value of our method, as its appeal lies beyond this particular application.

Indeed, a more detailed analysis shows our method's usefulness as a flagging system: it can point out significant blind spots of the Google baseline. This is remarkable in itself, considering the limited information it exploits (only the image modality, and no external training data). More precisely, we know the number $N_\mathrm{corr}$ characters on which our method corrects the baseline, of which $N_\mathrm{true}$ are actual corrections and $N_\mathrm{false}$ are false corrections. The following is a heuristic interpretation of $\Delta\mathrm{CER}$:
\begin{equation}
    \frac{-N_\mathrm{true}+N_\mathrm{false}}{N} \approx \Delta\mathrm{CER} \quad(\,\coloneqq \mathrm{CER}(\mathrm{base}+\mathrm{ours})-\mathrm{CER}(\mathrm{base}) \,) \,,
\end{equation}
where $N$ is the total number of characters in the ground truth. Together with $N_\mathrm{corr}=N_\mathrm{true}+N_\mathrm{false}$, we can then estimate the \emph{correction accuracy}
\begin{equation} \label{eq:acc}
    \mathrm{Acc} \coloneqq \frac{N_\mathrm{true}}{N_\mathrm{corr}} \approx \frac{1}{2}\left(1-\frac{\Delta\mathrm{CER}}{N_\mathrm{corr}/N}\right)
\end{equation}
which we show in \Cref{subtab:acc}. It appears that even when applied on Google, our method has statistically significant accuracy, so flagged errors can then be usefully corrected downstream, either by a human or a language model. This demonstrates that \emph{there is fundamentally untapped room for improvement in OCR}, since even state-of-the art systems under-use the internal redundancy of their inputs, while they rely on training data and language modeling (see \cref{sec:relatedwork}): our method, using the former but foregoing the latter two, was able to highlight a statistically significant number of recognition errors of those systems.

\begin{table}
    \centering
    \subfloat[]{\label{subtab:cer}
        \begin{tabular}{@{}lc|c@{}}
        \toprule
        & Berrutti & ENP \\
        \midrule
        CharNet & 33.2$\pm$0.5 & 28.2$\pm$0.5 \\
        + ours & $-$1.16$\pm$0.06 & $-$1.30$\pm$0.06 \\
        \midrule
        Google & 6.1$\pm$0.4 & 3.3$\pm$0.2 \\
        + ours & $+$0.53$\pm$0.05 & $+$0.23$\pm$0.04 \\
        \bottomrule
        \end{tabular}
    }
    \subfloat[]{\label{subtab:acc}
        \begin{tabular}{@{}lc|c@{}}
        \toprule
        & Berrutti & ENP \\
        \midrule
        CharNet + ours & 78$\pm$1.4 & 74$\pm$1.2 \\
        \midrule
        Google + ours & 34$\pm$2.0 & 31$\pm$3.8 \\
        \bottomrule
        \end{tabular}
    }
    \caption{\protect\subref{subtab:cer} CER of the base OCR (resp.\ $\Delta$CER from the base to our correction, where negative values indicate improvement), in \%, computed on each sub-collection (as average over lines weighted by line length), then aggregated by dataset (with unweighted average). Shown values indicate the 95\% confidence interval obtained by bootstrapping, sub-collection-wise, over $10^4$ resamples. \protect\subref{subtab:acc} Correction accuracy, in \%, as defined in \Cref{eq:acc} (heuristic estimation).}
    \label{tab:metrics}
\end{table}

\paragraph{Ablation study}

\begin{figure*}
\centering
\begin{tabular}{ccc}
     & Berrutti & ENP \\
    \rotatebox[origin=c]{90}{CharNet}
        & \parbox[c]{0.48\columnwidth}{\includeinkscape[width=0.48\columnwidth]{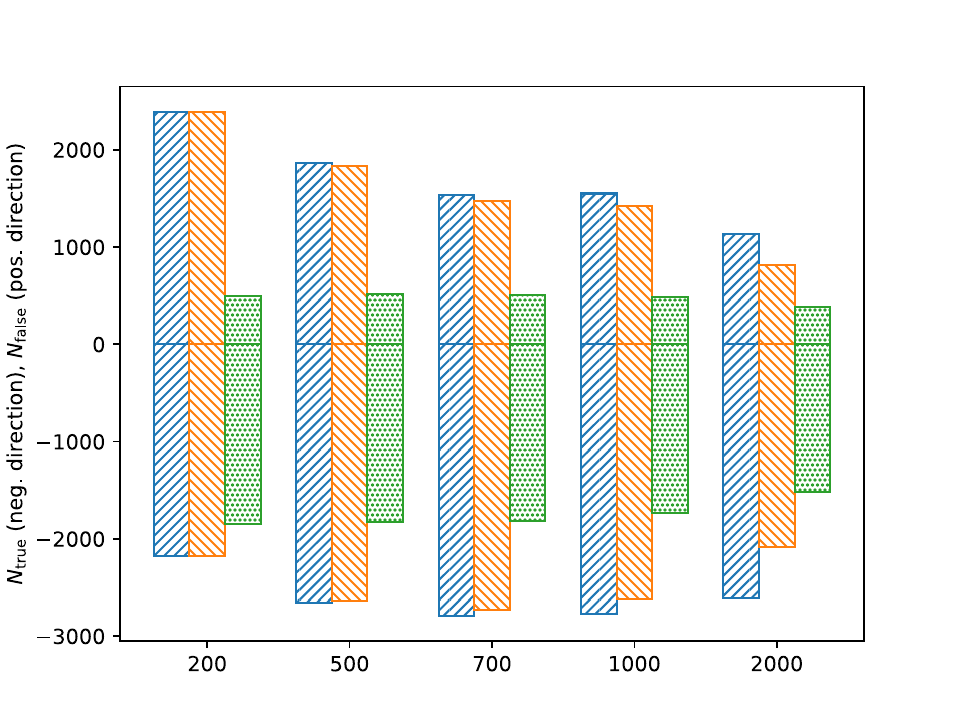}}
        & \parbox[c]{0.48\columnwidth}{\includeinkscape[width=0.48\columnwidth]{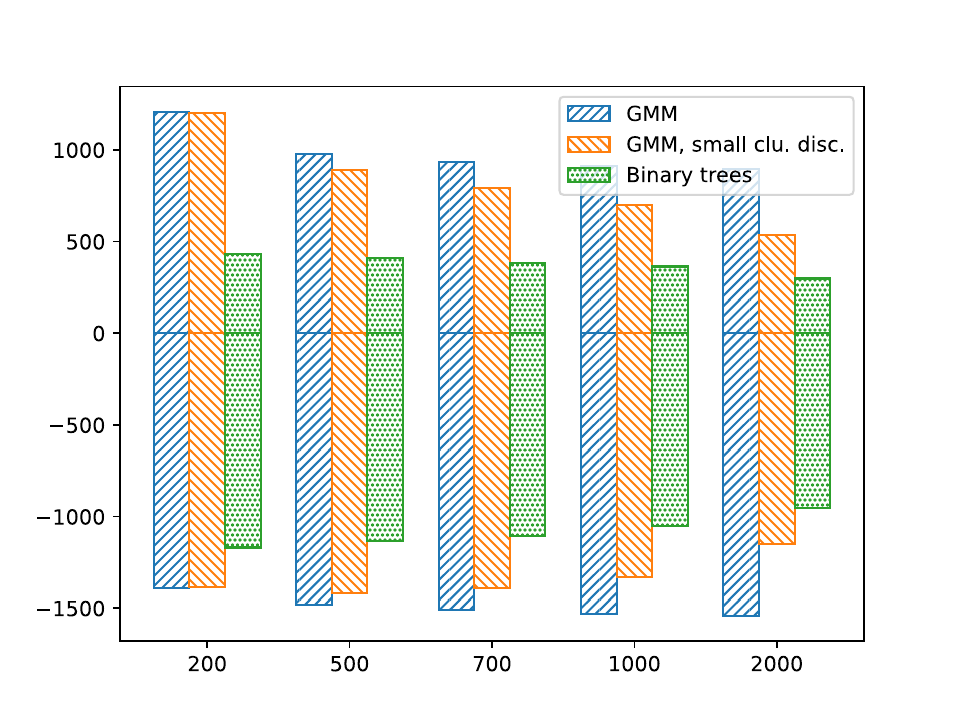}} \\
    \rotatebox[origin=c]{90}{Google}
        & \parbox[c]{0.48\columnwidth}{\includeinkscape[width=0.48\columnwidth]{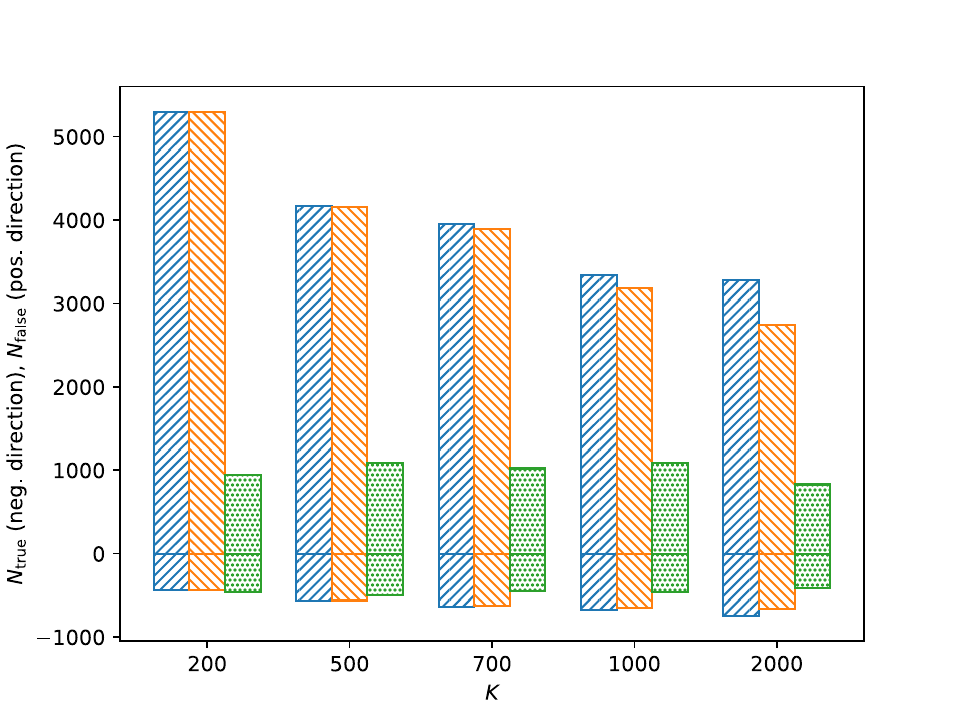}}
        & \parbox[c]{0.48\columnwidth}{\includeinkscape[width=0.48\columnwidth]{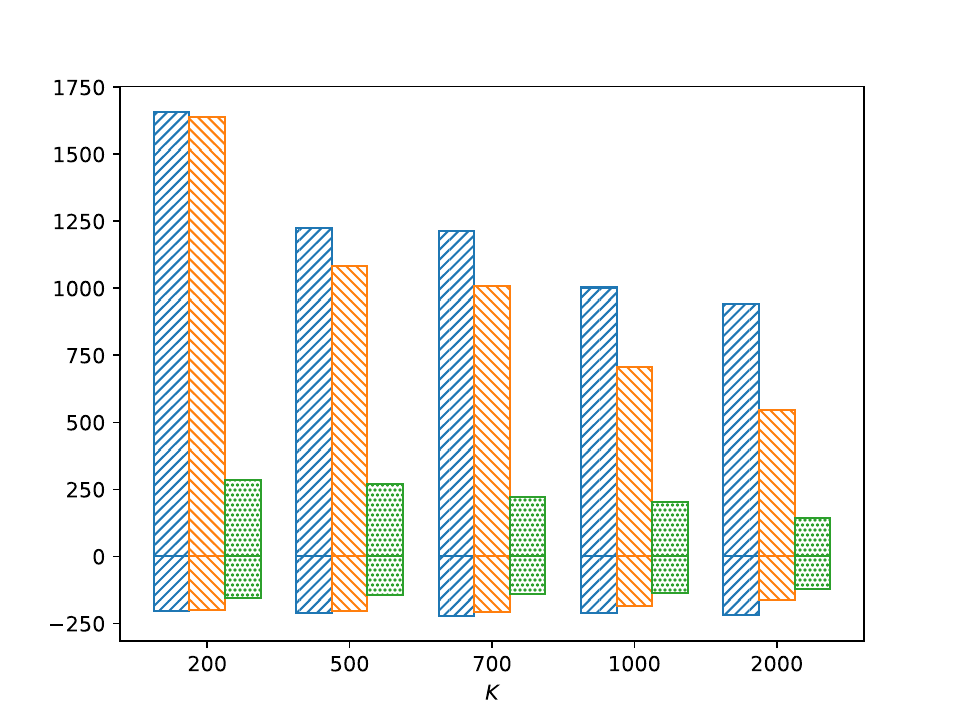}} \\
\end{tabular}
\caption{$N_\mathrm{false}$ (resp. $-N_\mathrm{true}$) on the positive (resp. negative) scale, heuristically estimated on each
sub-collection as in \Cref{eq:acc}, then averaged
by dataset (unweighted). The algorithm is run for different values of $K$, in one of three ablated versions: using the clustering output of the GMM step, using the same output from which small clusters (fewer than $n_\mathrm{min}$ elements) are discarded, and using the output of the binary tree refinement (\ie unablated).}
\label{fig:ablation_gmm}
\end{figure*}

We investigate the specific effect of the binary tree cluster refinement module of \Cref{ssec:trees}, which is the main novelty of our method. To this end, we ablate it from the pipeline, by running the final OCR error correction module of \Cref{ssec:relabeling} directly on the results of the GMM fitting step of \Cref{ssec:GMM}. For a fair comparison, we note that the refinement module's main action is to \emph{create} new clusters by splitting existing ones, but it also \emph{discards} clusters of insignificant size; therefore we also test a variant of the GMM clustering by similarly discarding the small clusters of the output. All experiments are done across various values for the starting number $K$ components in the GMM. Results are shown in \Cref{fig:ablation_gmm} as the numbers of true (resp. false) corrections $N_\mathrm{true}$ (resp. $N_\mathrm{false})$ as defined in \Cref{eq:acc}, rather than as $\Delta\mathrm{CER}$, so that we can precisely evaluate the impact of ablated parts. Two observations can be drawn from those results, demonstrating the \emph{performance} and the \emph{robustness} of our cluster refinement.

First, the binary tree refinement step has a consistent added value in performance across the board: while it tends to reduce both $N_\mathrm{false}$ and $N_\mathrm{true}$, the former is reduced more than the latter. In other words, the refinement step suggests fewer corrections but keeps more accurate corrections than false ones for an overall positive impact. More precisely, the reduction is more significant for $N_\mathrm{false}$ than for $N_\mathrm{true}$ in absolute difference, so that the CER is improved, as well as in relative factor, so that the correction accuracy is improved. Note that the scale of the impact depends on the base OCR, and that the most significant improvements come in the most challenging cases (Google).

\begin{table*}
\centering
\subfloat[]{ \label{subtab:numclu}
    \resizebox{0.48\columnwidth}{!}{
    \begin{tabular}{@{}ll|ccccc|ccccc@{}}
    \toprule
    & Dataset & \multicolumn{5}{|c}{Berrutti} & \multicolumn{5}{|c}{ENP}\\
    & $K$ & 200 & 500 & 700 & 1000 & 2000 & 200 & 500 & 700 & 1000 & 2000\\
    \midrule
    & GMM & 199 & 499 & 697 & 992 & 1941 & 199 & 492 & 681 & 952 & 1778 \\
    & GMM, small clu. disc. & 198 & 460 & 568 & 674 & 806 & 181 & 295 & 316 & 320 & 305 \\
    & Binary trees & 1276 & 1287 & 1313 & 1289 & 1233 & 581 & 583 & 564 & 558 & 503 \\
    \midrule
    & GMM & 199 & 499 & 698 & 995 & 1960 & 199 & 491 & 678 & 949 & 1769 \\
    & GMM, small clu. disc. & 199 & 478 & 624 & 799 & 1022 & 178 & 283 & 305 & 314 & 285 \\
    & Binary trees & 1571 & 1606 & 1604 & 1611 & 1561 & 538 & 536 & 534 & 520 & 452 \\
    \bottomrule
    \end{tabular}}}
\subfloat[]{ \label{subtab:propchars}
    \resizebox{0.48\columnwidth}{!}{
    \begin{tabular}{@{}ll|ccccc|ccccc@{}}
    \toprule
    & Dataset & \multicolumn{5}{|c}{Berrutti} & \multicolumn{5}{|c}{ENP}\\
    & $K$ & 200 & 500 & 700 & 1000 & 2000 & 200 & 500 & 700 & 1000 & 2000\\
    \midrule
    & GMM, small clu. disc. & 100 & 100 & 98 & 96 & 88 & 99 & 93 & 89 & 84 & 70 \\
    & Binary trees & 75 & 75 & 74 & 73 & 68 & 71 & 69 & 67 & 63 & 56 \\
    \midrule
    & GMM, small clu. disc. & 100 & 100 & 99 & 98 & 92 & 99 & 92 & 88 & 82 & 67 \\
    & Binary trees & 74 & 74 & 75 & 74 & 70 & 74 & 72 & 70 & 66 & 56 \\
    \bottomrule
    \end{tabular}}}
\caption{Complementary data for the ablation study (see \cref{fig:ablation_gmm}). \protect\subref{subtab:numclu} Number of clusters. \protect\subref{subtab:propchars} Proportion, in \%, of characters in retained (non-discarded) clusters.}
\label{tab:ablation_gmm_numclusters}
\end{table*}

Second, while the performance of the GMM clustering greatly depends on the parameter $K$, the downstream performance after cluster refinement does not (except for very large $K$). This step has a remarkable \emph{stabilizing effect}. This property is further confirmed when looking at the number of clusters created by the method, and the proportion of characters included in retained clusters, shown in \Cref{tab:ablation_gmm_numclusters}: whatever the GMM clustering results, the refinement algorithm will split those in a way that leads to a consistent result. This phenomenon starts to break down when $K$ gets too large. This can be explained by most clusters being too small, leading the refinement algorithm to a regime where cluster discarding is predominant. Otherwise, one can conclude that the parameter $K$ can be chosen relatively freely without affecting the quality of the pipeline output, the refinement step being robust enough.

\subsection{Qualitative analysis} \label{ssec:qual}

\begin{figure}[t]
    \centering
    \subfloat[]{\label{subfig:successes}
        \begin{minipage}{0.48\columnwidth}
        \centering
        \includegraphics[width=0.9\columnwidth]{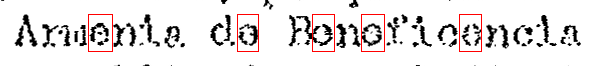} \\
        \includegraphics[width=0.9\columnwidth]{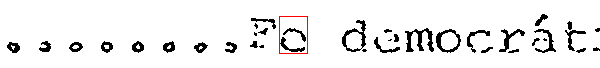} \\
        \includegraphics[width=0.9\columnwidth]{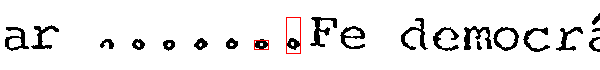} \\
        \includegraphics[width=0.9\columnwidth]{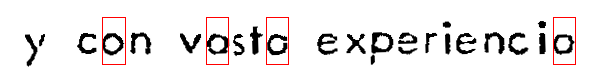} \\
        \includegraphics[width=0.9\columnwidth]{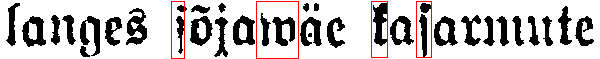}
        \end{minipage}
    }
    \subfloat[]{\label{subfig:failures}
        \begin{minipage}{0.48\columnwidth}
        \centering
        \includegraphics[width=0.9\columnwidth]{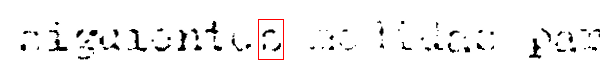} \\
        \includegraphics[width=0.9\columnwidth]{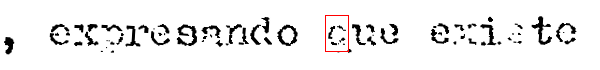} \\
        \includegraphics[width=0.9\columnwidth]{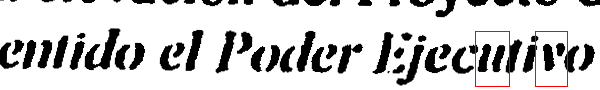} \\
        \includegraphics[width=0.9\columnwidth]{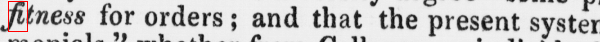} \\
        \includegraphics[width=0.9\columnwidth]{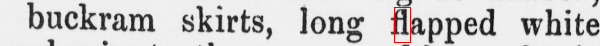} \\
        \includegraphics[width=0.9\columnwidth]{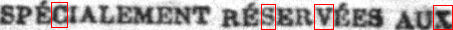} \\
        \includegraphics[width=0.9\columnwidth]{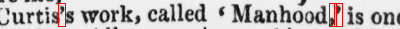} \\
        \end{minipage}
    }
    \caption{Examples of character detections by Google. The predicted labels of drawn bounding boxes are the following: `Google label' $\rightarrow$ `our correction' (with boldface indicating agreement with ground truth). \protect\subref{subfig:successes} Correct relabels. Line~1: `\textbf{e}'$\rightarrow$`\textbf{e}', `o'$\rightarrow$`\textbf{e}', `o'$\rightarrow$`\textbf{e}', `o'$\rightarrow$`\textbf{e}', `\textbf{e}'$\rightarrow$`\textbf{e}'. Line~2: `o'$\rightarrow$`\textbf{e}'. Line~3: `°'$\rightarrow$`\textbf{.}', `\textbf{.}'$\rightarrow$`\textbf{.}'. Line~4: `\textbf{o}'$\rightarrow$`\textbf{o}', `o'$\rightarrow$`\textbf{a}', `o'$\rightarrow$`\textbf{a}', `o'$\rightarrow$`\textbf{a}'. Line~5: `j'$\rightarrow$`\textbf{s}', `m'$\rightarrow$`\textbf{w}', `f'$\rightarrow$`\textbf{k}', `j'$\rightarrow$`\textbf{s}'. \protect\subref{subfig:failures} Wrong relabels. Line~1: `\textbf{s}'$\rightarrow$`o'. Line 2: `\textbf{q}'$\rightarrow$`e'. Line 3: `\textbf{u}'$\rightarrow$`n', `\textbf{v}'$\rightarrow$`r'. Line 4: `i'$\rightarrow$`h'. Line 5: `l'$\rightarrow$`i'. Line 6: `\textbf{C}'$\rightarrow$`c', `\textbf{S}'$\rightarrow$`s', `\textbf{V}'$\rightarrow$`v', `\textbf{X}'$\rightarrow$`x'. Line 7: `\textbf{'}'$\rightarrow$`,', `\textbf{,}'$\rightarrow$`\textbf{,}', `\textbf{'}'$\rightarrow$`,'.}
    \label{fig:outputexamples} 
\end{figure}

In \Cref{fig:outputexamples} we present notable examples of our method's success and failure cases. They show corrections made by our method on the Google baseline across different dataset documents. We do not claim that these examples represent all the results; instead, they have been selected because they showcase a few phenomena we have observed and thus provide some practical understanding of our method.

In \Cref{subfig:successes}, almost all the highlighted characters have been incorrectly labeled by Google but successfully corrected by our post-processing. We deliberately chose difficult examples: lines 1 and 2 show heavily degraded images. In contrast, lines 2 to 5 display unusual glyphs, particularly lines 4 and 5, using fonts (historical or contemporary) where some glyphs resemble the shape of different symbols in more common fonts. In those cases, the character recognition task is impossible without using context or \emph{document-specific prior knowledge}. Our method can handle them by acquiring the document-specific knowledge of which characters must be clustered together, \ie the \emph{redundancy} of each glyph's occurrence across the document. A detailed inspection of the provided example reveals that they are part of uniform clusters, as desired. Note how the subtle difference between `o' and `a' in line 4 is correctly discriminated. At the same time, our method's accurate labeling of the clusters demonstrates that the base OCR system could correctly recognize most, but not all, of the elements of each cluster. This inconsistent success of the base OCR (here, the SOTA Google system) hints at its leveraging of the local textual context or the breadth of its general prior knowledge. Still, it is an empirical proof that it does not use the global context within the document.

Conversely, \Cref{subfig:failures} is an overview of the typical situations where our method fails. Lines 1 and 2 show characters whose image is degraded to become unrecognizable. In those cases, our method tends to aggregate images of different glyphs into the same cluster, which passes all normality tests, probably because the degradations provide more variance than shape differences between glyphs. The highlighted characters in lines 3 to 5 have few occurrences across the document collection. Our method puts them in the wrong cluster, where they are outliers; a visual inspection nonetheless shows that they do look similar to the model of their attributed cluster (in particular, the italics `ﬁ' ligature is in a cluster of italics `h', and the `ﬂ' ligature is in a cluster of `ﬁ' which are labeled as `i' by the baseline OCR). Finally, lines 6 and 7 show how our clustering algorithm only see aligned cropped images of characters, hence has no clue about their size or position: uppercase letters can then look identical to their lowercase version, as can apostrophes look identical to commas. Note that these are cases where clustering step is performing correctly.

\section{Conclusion}\label{sec:conclusion}

This work explores the following hypothesis on the current state of automatic printed text recognition systems: they rely on geometric patterns identified through huge amounts of training data or language patterns, but under-use the patterns of the geometric redundancy of glyphs internal to any given input document. We test this by developing a method to correct character recognition errors in post-processing, relying solely on internal document geometric redundancy without prior training. Experiments on various data (including a dataset we introduce here, the Berrutti OCR Collection) show that our method, while yet impractical for fully automated post-processing, demonstrates aptitude as a flagging system to detect errors with statistically significant accuracy, hence validating our hypothesis.

\paragraph{Future perspectives.}

Even as our method's core unsupervised clustering algorithm exhibits robustness and stability properties, experiments point to specific directions to further improve its performance, \eg by incorporating positional information. Beyond this, our algorithm provides valuable modeling information that opens up new downstream possibilities. For example, while our paper focuses on correcting substitution errors, a combined use of our character models with pattern matching techniques could tackle insertion and deletion errors. In another possible application, post-OCR systems based on stochastic language models can incorporate accurate confidence estimation derived from our model.

\begin{credits}
\subsubsection{\ackname} 
The research that originated the results presented in this
publication was partly supported by the Agencia Nacional de Investigación e Innovación under scholarships POS\_EXT\_2023\_2\_180123 and POS\_NAC\_2024\_1\_183502, and by the France 2030 CollabNext project. The experiments presented in this paper used ClusterUY \cite{clusterUy}. Centre Borelli is also a member of Université Paris Cité, CNRS, SSA and INSERM.
\end{credits}

\bibliographystyle{splncs04}
\bibliography{bibliography}

\appendix

\section*{Supplementary material}

This supplementary material includes additional quantitative results and details on our methods. \Cref{app-sec:method_details} provides a precise description of our method's preprocessing step, as wee as pseudo-code implementations for several key steps. \Cref{app-sec:detailed_quant} provides all quantitative experimental results at the sub-collection granularity, where the main paper only shows numbers aggregated by dataset.

\section{Our method: thorough description and pseudo-code implementations} \label{app-sec:method_details}

This section is complementary to \Cref{sec:clustering}, providing further explanations of our method for increased reproducibility. \Cref{app-ssec:preprocessing_radio,app-ssec:preprocessing_geom} describe our image and bounding-box pre-processing steps respectively. \Cref{app-ssec:algorithms} contains pseudo-code implementations of several of our proposed algorithms, including the aforementioned pre-processing as well as cluster refinement and final character relabeling.

\subsection{Poisson image standardization} \label{app-ssec:preprocessing_radio}

\begin{figure}
    \centering
    \subfloat[Document extract (from the polish sub-collection of ENP).]{\label{app-subfig:radio_doc}
        \begin{minipage}{0.48\columnwidth}
        \centering
        \includegraphics[width=\columnwidth]{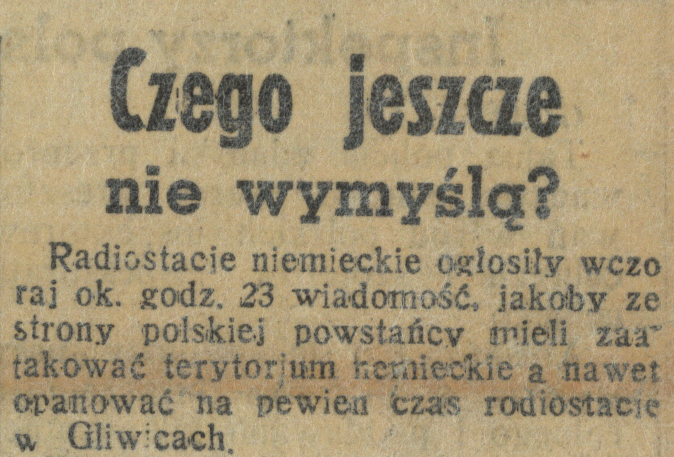} \\
        \end{minipage}
    } \\
    \subfloat[Inputs: lines (manual segmentation).]{\label{app-subfig:radio_input}
        \begin{minipage}{0.48\columnwidth}
        \centering
        \includegraphics[width=\columnwidth]{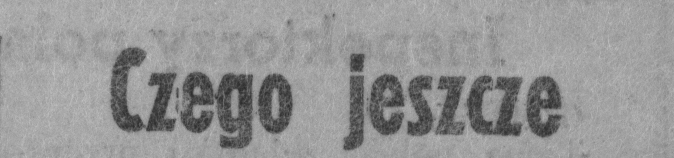} \\
        \includegraphics[width=\columnwidth]{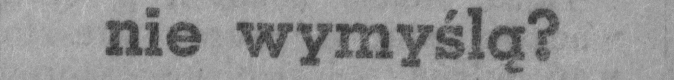} \\
        \includegraphics[width=\columnwidth]{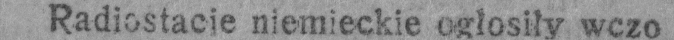} \\
        \includegraphics[width=\columnwidth]{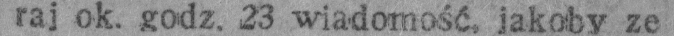} \\
        \includegraphics[width=\columnwidth]{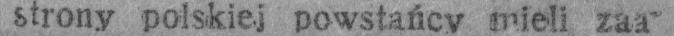} \\
        \includegraphics[width=\columnwidth]{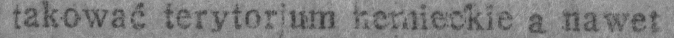} \\
        \includegraphics[width=\columnwidth]{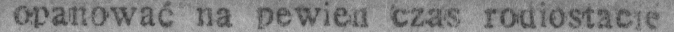} \\
        \includegraphics[width=\columnwidth]{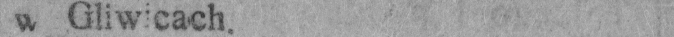} \\
        \end{minipage}
    }
    \subfloat[Foreground masks.]{\label{app-subfig:radio_mask}
        \begin{minipage}{0.48\columnwidth}
        \centering
        \includegraphics[width=\columnwidth]{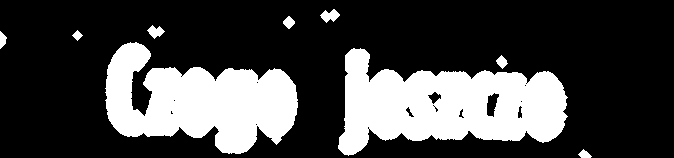} \\
        \includegraphics[width=\columnwidth]{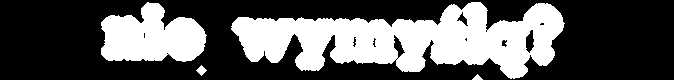} \\
        \includegraphics[width=\columnwidth]{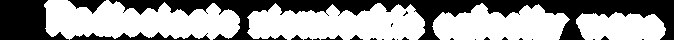} \\
        \includegraphics[width=\columnwidth]{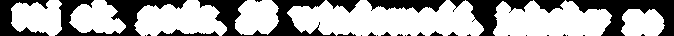} \\
        \includegraphics[width=\columnwidth]{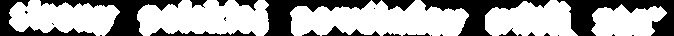} \\
        \includegraphics[width=\columnwidth]{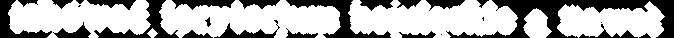} \\
        \includegraphics[width=\columnwidth]{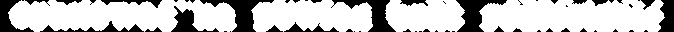} \\
        \includegraphics[width=\columnwidth]{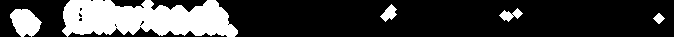} \\
        \end{minipage}
    } \\
    \subfloat[Poisson editing results.]{\label{app-subfig:radio_poisson}
        \begin{minipage}{0.48\columnwidth}
        \centering
        \includegraphics[width=\columnwidth]{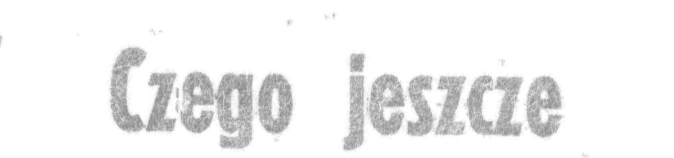} \\
        \includegraphics[width=\columnwidth]{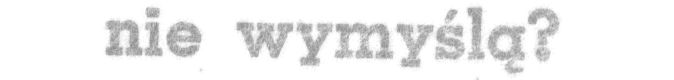} \\
        \includegraphics[width=\columnwidth]{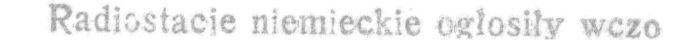} \\
        \includegraphics[width=\columnwidth]{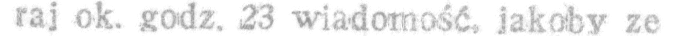} \\
        \includegraphics[width=\columnwidth]{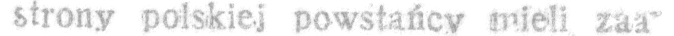} \\
        \includegraphics[width=\columnwidth]{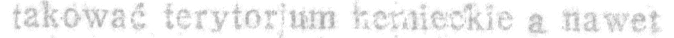} \\
        \includegraphics[width=\columnwidth]{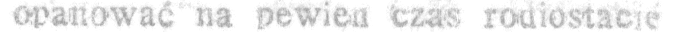} \\
        \includegraphics[width=\columnwidth]{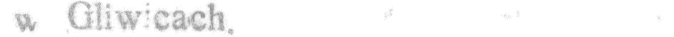} \\
        \end{minipage}
    }
    \subfloat[Contrast-adjusted outputs]{\label{app-subfig:radio_final}
        \begin{minipage}{0.48\columnwidth}
        \centering
        \includegraphics[width=\columnwidth]{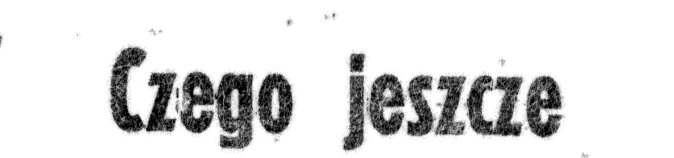} \\
        \includegraphics[width=\columnwidth]{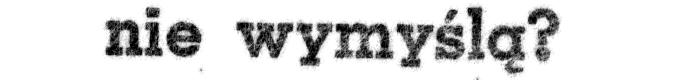} \\
        \includegraphics[width=\columnwidth]{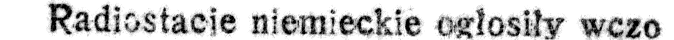} \\
        \includegraphics[width=\columnwidth]{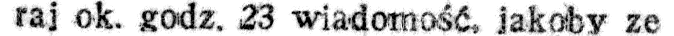} \\
        \includegraphics[width=\columnwidth]{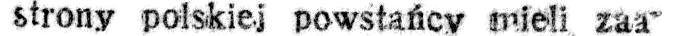} \\
        \includegraphics[width=\columnwidth]{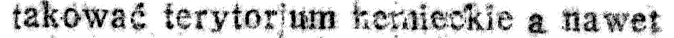} \\
        \includegraphics[width=\columnwidth]{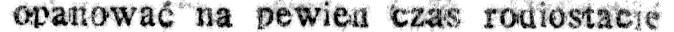} \\
        \includegraphics[width=\columnwidth]{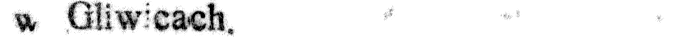} \\
        \end{minipage}
    }
    \caption{Our Poisson image standardization process illustrated on a few examples. Note how noise, color variations and show-through text are eliminated from the background.}
    \label{app-fig:examples_radio} 
\end{figure}

At the very beginning of our method, radiometric pre-processing of line images is performed to make the background uniformly white. Consequently, character images can be cropped and padded for downstream steps, without issues of image discontinuity at the crop borders. The process given in \Cref{app-alg:poisson} and illustrated in \Cref{app-fig:examples_radio}. In the following, we assume images to be gray-scale; if not, the three color channels are averaged (\cref{app-subfig:radio_input}).

First, a mask encompassing all foreground (text) pixels is created by Otsu thresholding and morphological dilation (\cref{app-subfig:radio_mask}). Dilation is carried out repeatedly (five times in our implementation): the masks does not need to be precise, we find that including background pixels is preferable to missing foreground pixels.

Then, the image is masked and seamlessly edited onto a white background by Poisson editing (\cref{app-subfig:radio_poisson}). More precisely, let the image grid be a graph $(V, E)$ where vertices $v\in V$ are pixels and edges $e=(v_1,v_2)\in E$ are pairs of neighboring pixels (with 4-connectivity, so that interior, border and corner pixels have degree 4, 3 and 2 respectively); edges are oriented so as to flow to the right or to the bottom (positive $x$- and $y$-direction). An image (here, of a text line) is a vector $u\in[0, 1]^V$ (with values 0 for blacks and 1 for whites); the mask computed above defines a subset $\Omega\subset V$ of vertices. One has the standard linear operators of gradient $\nabla\colon\R^V\to\R^E$, divergence $\divergence\colon\R^E\to\R^V$ and Laplacian $\Delta\colon\R^V\to\R^V$; they are represented by the matrices $B$, $-B^\top$ and $-B^\top B$ respectively, where $B\in M_{V,E}(\R)$ is the incidence matrix of the graph.

We define a target gradient field $\mathbf{g}\in\R^E$ by imposing it to be $\nabla u$ in the foreground and $0$ elsewhere:
\begin{equation} \label{app-eq:target_gradient}
    \forall e\in E, \quad \mathbf{g}(e) =
    \begin{cases}
        \nabla u(e) & \text{if }e\in \Omega^2 \\
        0 & \text{otherwise.}
    \end{cases}
\end{equation}
The desired edited image $v\in\R^V$ is then the solution of the following Poisson equation:
\begin{align} \label{app-eq:poisson}
    \left\{\begin{aligned}
        \Delta v &= \divergence(\mathbf{g}) && \text{inside }\Omega \\
        v &= 1 && \text{outside }\Omega
    \end{aligned}\right.
\end{align}
which is an invertible linear system in $\R^V$. Note that formulating the problem on the graph $(V, E)$, instead of choosing discrete differential schemes, seamlessly handles the boundary conditions, using Dirichlet conditions at the boundary $\Omega$ except where it meets the edges of the images, where Neumann conditions are used instead.

Finally, we change the contrast to keep black levels consistent (\cref{app-subfig:radio_final}). Indeed, Poisson editing keeps invariant the intensity difference between foreground and background: given two input images $u$ and $u'$ whose foregrounds' blacks have the same intensity, if the background's white is grayer for $u$ than for $u'$, then in the Poisson-edited outputs, the blacks will be less intense in $v$ than in $v'$. We remedy this by applying a linear contrast change, mapping to $1$ the white intensity $1$, and to $0.1$ the first decile of non-white intensity values (\ie strictly lower than $1$). Last, the image is clipped to keep intensity values inside $[0, 1]$.

\subsection{Character bounding-box pre-processing} \label{app-ssec:preprocessing_geom}

\begin{figure}
    \centering
    \subfloat[Input line image (from the english sub-collection of ENP).]{\label{app-subfig:geom_input}
        \includegraphics[width=0.96\columnwidth]{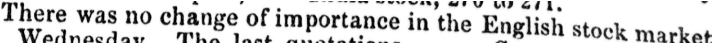}
    } \\
    \subfloat[Proto-lines 1/3: horizontal blur.]{\label{app-subfig:geom_hblur}
        \includegraphics[width=0.96\columnwidth]{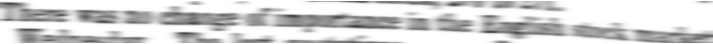}
    } \\
    \subfloat[Proto-lines 2/3: binarized result.]{\label{app-subfig:geom_binary}
        \includegraphics[width=0.96\columnwidth]{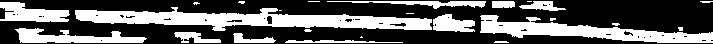}
    } \\
    \subfloat[Proto-lines 3/3: grouping of character boxes (see color version of the paper).]{\label{app-subfig:geom_boxgrouping}
        \includegraphics[width=0.96\columnwidth]{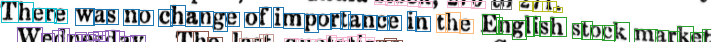}
    } \\
    \subfloat[Cropped proto-line.]{\label{app-subfig:geom_protoline}
        \includegraphics[width=0.96\columnwidth]{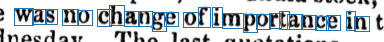}
    } \\
    \subfloat[Guidance cost map for horizontal paths: dark purple is 0, bright yellow is 0.3.]{\label{app-subfig:geom_guidance}
        \includegraphics[width=0.96\columnwidth]{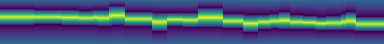}
    } \\
    \subfloat[Horizontal paths: upper and lower text boundaries.]{\label{app-subfig:geom_hpaths}
        \includegraphics[width=0.96\columnwidth]{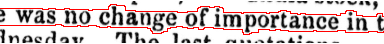}
    } \\
    \subfloat[Final result 1/2: output masks of every other character (for visual clarity).]{\label{app-subfig:geom_output1}
        \includegraphics[width=0.96\columnwidth]{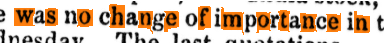}
    } \\
    \subfloat[Final result 2/2: output masks of every other character (for visual clarity).]{\label{app-subfig:geom_output2}
        \includegraphics[width=0.96\columnwidth]{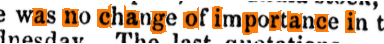}
    }
    \caption{Our bounding-box pre-processing pipeline illustrated on an example. The boxes predicted by the base OCR produce an imperfect segmentation of characters (Fig.\ \protect\subref{app-subfig:geom_protoline}), which our masks correct (Figs.\ \protect\subref{app-subfig:geom_output1} and \protect\subref{app-subfig:geom_output2}).}
    \label{app-fig:examples_geom} 
\end{figure}

The other side of our pre-processing concerns the geometry of the detection data: starting from the character bounding boxes produced by the base OCR, we produce fine segmentation masks for precise cropping. We propose a procedure relying on elementary algorithmic bricks, first drawing the upper and lower horizontal boundaries of the text line, then finding the vertical lines separating the characters from one another. It is illustrated in \Cref{app-fig:examples_geom}.

Before drawing boundary lines, character boxes are have to be clustered into ``proto-lines'', which are groups of consecutive characters along the same line. The reason for doing this is that input line images sometimes include parts of the previous or following line in the source document, from which the base OCR can detect characters; therefore it would be ideal to segment those detections into lines. We have not achieved a way to reliably do so, hence we relax the problem by allowing over-segmentation, and proceed in the following way. Proto-lines are defined as the connected components of a special binarization of the image. To have the longest possible proto-lines, while keeping different lines from merging, we perform this binarization as such: a 1-D gaussian filter is applied (\cref{app-subfig:geom_hblur}), then Otsu thresholding is independently performed on each vertical slice\footnote{We avoid doing a global Otsu thresholding as we find it to poorly handle cases where text stroke have non-uniform width: bolder words tend to invisibilize the others.} (\cref{app-subfig:geom_binary}). With the proto-lines defined as mask components, each character bounding box is matched to the proto-line having the biggest intersection (\cref{app-subfig:geom_boxgrouping}). The rest of the process runs on each proto-line, which are horizontally cropped (\cref{app-subfig:geom_protoline}).

Horizontal upper and lower text boundaries are computed as paths (\ie sequences of adjacent pixels) minimizing the total cost, for a given cost map (the cost of a path is the sum of individual pixel costs). The cost map is the inverted image (0 for whites and 1 for blacks) to which we add a regularizing ``guidance'' map. The guidance map is built to push the paths towards the input bounding boxes: it is 0 at the top (resp. bottom) edges of the boxes, it outwardly linearly increases to a chosen value $\lambda_h=0.1$ at the top (resp. bottom) edge of the image, and inwardly linearly increases to $3\lambda_h$ at the halfway points between top and bottom of the boxes (\cref{app-subfig:geom_guidance}). Last, the cost map is set to infinity along a horizontal barrier running through the middle of the boxes, to prevent paths from crossing the text. Given this cost map, the minimizing path, among those running from the top-left-most box corner to the top-right-most box corner, is chosen as the upper boundary of all character segmentation masks for the proto-line (and analogously for the lower boundary); see \cref{app-subfig:geom_hpaths}.

Then, for each character box of the proto-line, vertical (left and right) edges are similarly replaced by a minimizing path: they run from the intersection of the edge and the upper boundary, to that of the edge and lower boundary. As before, the cost map is taken as the sum of the inverted image and of a regularizer map; this time however the regularizer is uniformly set to a value $\lambda_v$\footnote{Contrary to the horizontal case, prior information from the boxes is not used to build the regularizer map, for two reasons. First, inter-character space is narrower than inter-line space, so paths are already more constrained; additional constraints risk overriding too many of the original constraints. Second, the vertical edges predicted by the base OCRs tend to be less reliable.}. This value sets the rigidity of the path: a lower $\lambda_v$ allows the path to take bigger detours, a higher $\lambda_v$ forces it to be straighter. The correct value is highly dependent on the document. Therefore we set it iteratively: the left and right paths are first computed with $\lambda_v=0.05$, and if they are not acceptable, the process is repeated with $\lambda_v$ incremented by $\delta\lambda_v=0.15$; until acceptability, or a number of eight iterations, is reached. Acceptability is defined by the interval between both paths: we require it to be on average within $10\%$ of the box width, and that its biggest deviation be within $1/3$ of the box width. At each iteration, besides incrementing $\lambda_v$, we also shift the positions of the box edges, for exploration in case those edges were incorrectly predicted by the base OCR. The left edge $x=l$ of the current box $B$ is shifted by two pixels towards the closest right edge $x=r'$, among boxes $B'$ such that the left edge $x=l'$ of $B'$ satisfies $l'\leq l$ (and analogously for the right edge); this is based on the principle that if $B'$ is the box preceding $B$, then one should have $r'=l$ ``in ground truth''.

For each character, having computed the left and right vertical paths, we define the mask as the set of pixels lying between those two paths and the upper and lower boundaries of the corresponding proto-line (\cref{app-subfig:geom_output1,app-subfig:geom_output2}).

\subsection{Pseudo-codes} \label{app-ssec:algorithms}

For maximum clarity, we detail here implementations in pseudo-code of core parts of our method: \Cref{app-alg:poisson} is the radiometric pre-processing (see \cref{ssec:preprocessing,app-ssec:preprocessing_radio}), \Cref{app-alg:tree} is the cluster refinement step (\cref{ssec:trees}), and \Cref{app-alg:relabeling} is the final OCR correction step (\cref{ssec:relabeling}). Not included are pseudo-codes for the GMM clustering (\cref{ssec:GMM}), which is explicit enough, and for the geometric pre-processing (\cref{app-ssec:preprocessing_geom}), which has too many minutiae; for the latter we refer the interested reader to our published source code.

\begin{algorithm}
    \caption{PoissonStandardization} 
    \label{app-alg:poisson}
    \DontPrintSemicolon
    
    \Input{$u$: image, in $[0,1]^V\simeq[0,1]^{H\times W}$}
    \Parameter{$d$: dilation factor}
    \Output{$v$: standardized image, in $[0,1]^V$}

    \tcp*[h]{Foreground mask} \;
    $t \leftarrow$ Otsu threshold of $u$ \;
    $\Omega \leftarrow \enstq{(i, j)\in V}{u(i, j)\leq t}$ \;
    \For{$k \leftarrow 1$ \KwTo $d$}{
        $\Omega \leftarrow$ binary dilation of $\Omega$ \;
    }
    \tcp*[h]{Poisson editing} \;
    $\mathbf{g} \leftarrow$ target gradient field defined in \Cref{app-eq:target_gradient} \;
    $v \leftarrow$ solution $v(u, \Omega, \mathbf{g})$ of \Cref{app-eq:poisson} \;
    \tcp*[h]{Contrast change and clipping} \;
    $q \leftarrow$ first decile of the sample $\enstq{v(i,j)}{(i,j)\in V,\,v(i,j)<1}$ \;
    $v(i,j) \leftarrow \max\left( \min\left( 1-0.9\frac{1-v(i,j)}{1-q}, 1 \right), 0 \right)$ for all $(i,j)\in V$ \;
    \Return{v}
\end{algorithm}

\begin{algorithm}
    \caption{BinaryTreeRefinement} 
    \label{app-alg:tree}
    \DontPrintSemicolon
    
    \Input{$u_{1:n}$: cluster $\{u_1,\ldots,u_n\} \in(\R^{H\times W})^n$ of images, of size $n$}
    \Parameter{$n_\mathrm{min}$: minimal output cluster size}
    \Parameter{$k$: number of principal components to undergo normality test}
    \Parameter{$p_\mathrm{thr}$: threshold on $p$-values}
    \Output{$L$: list of sub-clusters}
    
    \If{$n<n_\mathrm{min}$}{
        \Return{$[\,]$}
    }
    $\mu \leftarrow \sum_iu_i/n$ \;
    \For{$i \leftarrow 1$ \KwTo $k$}{
        $u_i \leftarrow \mathrm{ICA}(u_i, \mu)$ \tcp*[h]{Register $u_i$ to reference $\mu$ with the Inverse Compositional Algorithm (with homothety warp group)}\;   
    }
    \For{$j \leftarrow 1$ \KwTo $k$}{
        \tcp*[h]{Normality tests on $k$ principal components} \;
        $v_j\in R^{H\times W}\leftarrow$ direction of $j$-th PC of $u_{1:n}$ \;
        $p_j \leftarrow$ Anderson-Darling $p$-value of $(v_j\cdot u)_{1:n}$
    }
    \If(\tcp*[h]{Return self if normality tests passed}){$\min_j{p_j}\geq p_\mathrm{thr}$}{
        \Return{$[(u_{1:n})]$}
    }
    $u^{(1)}_{1:n_1}, u^{(2)}_{1:n_2} \leftarrow$ partition of $u_{1:n}$ according to $\mathrm{GMM}_2$ on $x^{(1:k)}_{1:n}$ \;
    \If(\tcp*[h]{Return self if trivial partition}){$n_1=0$ or $n_2=0$}{
        \Return{$[(u_{1:n})]$}
    }
    $L_1 \leftarrow \mathrm{BinaryTreeRefinement}(u^{(1)}_{1:n_1})$ \;
    $L_2 \leftarrow \mathrm{BinaryTreeRefinement}(u^{(2)}_{1:n_2})$ \;
    \Return{$L_1+L_2$} \;
\end{algorithm}

\begin{algorithm}
    \caption{CorrectingLabelsUsingClusters} 
    \label{app-alg:relabeling}
    \DontPrintSemicolon
    \Input{$u_{1:N}$: list of character images}
    \Input{$l_{1:N}$: list of OCR-predicted labels (symbols)}
    \Input{$C_{1:m}$: list of clusters (disjoint partition of \{1,\ldots,N\})}
    \Parameter{$f_\mathrm{thr}$: majority frequency threshold}
    \Output{$l'_{1:N}$: list of corrected labels}
    
    $l'_{1:N} \leftarrow l_{1:N}$ \;
    \For{$k\leftarrow 1$ \KwTo $m$}{
        $l, f \leftarrow \argmax_l{\frac{\left\lvert\{i\in C_k\mid l_i = l\}\right\rvert}{\left\lvert C_k\right\rvert}}, \max_l{\frac{\left\lvert\{i\in C_k\mid l_i = l\}\right\rvert}{\left\lvert C_k\right\rvert}}$ \;
        \If{$f>f_\mathrm{thr}$}{
            $l'_{C_k} = (l \mid i\in C_k)$
        }
    \Return{$l'_{1:N}$}
    }
\end{algorithm}

\section{Detailed quantitative results} \label{app-sec:detailed_quant}

In this section, we expand on the quantitative experiments of \Cref{sec:experiments}, providing results for each sub-collection.

\subsection{Performance metrics: $\Delta$CER and correction accuracy}

\Cref{app-tab:metrics_berrutti,app-tab:metrics_enp} provide the detailed metrics presented in \Cref{tab:metrics}, broken down by sub-collection. They support the conclusions drawn in the main paper; we nonetheless note that our method is able to improve on Google in two sub-collections (ENP estonian and latvian), for which the latter is shown to struggle: see the last example of \Cref{subfig:successes}.

\begin{table}
    \centering
    \subfloat[]{\label{app-subtab:cer_berrutti}
        \begin{tabular}{@{}lc|c|c|c@{}}
            \toprule
            & high & med. & low & all \\
            \midrule
            CharNet & 16.8$\pm$0.6 & 32.0$\pm$0.8 & 50.7$\pm$1.3 & 33.2$\pm$0.5 \\
            + ours & $-$0.54$\pm$0.06 & $-$1.57$\pm$0.10 & $-$1.37$\pm$0.11 & $-$1.16$\pm$0.06 \\
            \midrule
            Google & 3.9$\pm$0.4 & 5.5$\pm$0.6 & 9.0$\pm$1.1 & 6.1$\pm$0.4 \\
            + ours & $-$0.01$\pm$0.05 & $+$0.49$\pm$0.07 & $+$1.12$\pm$0.12 & $+$0.53$\pm$0.05 \\
            \bottomrule
        \end{tabular}
    } \\
    \subfloat[]{\label{app-subtab:acc_berrutti}
        \begin{tabular}{@{}lc|c|c|c@{}}
            \toprule
            & high & med. & low & all \\
            \midrule
            CharNet + ours & 77$\pm$3.1 & 81$\pm$2.1 & 76$\pm$2.1 & 78$\pm$1.4 \\
            \midrule
            Google + ours & 51$\pm$4.2 & 26$\pm$3.6 & 27$\pm$2.5 & 34$\pm$2.0 \\
            \bottomrule
        \end{tabular}
    }
    \caption{\protect\subref{app-subtab:cer_berrutti} CER of the base OCR (resp.\ $\Delta$CER from the base to our correction, where negative values indicate improvement), in \%, computed on each sub-collection of Berrutti (as average over lines weighted by line length), then averaged (unweighted). Shown values indicate the 95\% confidence interval obtained by bootstrapping, sub-collection-wise, over $10^4$ resamples. \protect\subref{app-subtab:acc_berrutti} Correction accuracy, in \%, as defined in \Cref{eq:acc} (heuristic estimation).}
    \label{app-tab:metrics_berrutti}
\end{table}

\begin{table}
    \centering
    \subfloat[]{\label{app-subtab:cer_enp}
        \resizebox{0.95\columnwidth}{!}{
        \begin{tabular}{@{}lc|c|c|c|c|c|c|c|c|c@{}}
            \toprule
            & dut. & eng. & est. & fin. & fre. & ger. & lat. & pol. & swe. & all \\
            \midrule
            CharNet & 13.4$\pm$1.3 & 26.9$\pm$1.6 & 38.8$\pm$1.0 & 32.5$\pm$2.6 & 21.2$\pm$0.8 & 48.7$\pm$3.0 & 20.5$\pm$1.2 & 31.8$\pm$0.8 & 20.0$\pm$0.7 & 28.2$\pm$0.5 \\
            + ours & $-$0.86$\pm$0.16 & $-$1.02$\pm$0.11 & $-$2.40$\pm$0.15 & $-$1.50$\pm$0.26 & $-$1.71$\pm$0.13 & $-$1.56$\pm$0.29 & $-$0.67$\pm$0.13 & $-$1.64$\pm$0.13 & $-$0.38$\pm$0.10 & $-$1.30$\pm$0.06 \\
            \midrule
            Google & 4.0$\pm$0.7 & 2.0$\pm$0.2 & 4.0$\pm$0.3 & 3.3$\pm$0.9 & 1.3$\pm$0.2 & 3.7$\pm$0.8 & 7.5$\pm$0.8 & 2.9$\pm$0.3 & 1.5$\pm$0.3 & 3.3$\pm$0.2 \\
            + ours & $+$0.18$\pm$0.07 & $+$0.14$\pm$0.04 & $-$0.32$\pm$0.10 & $+$0.20$\pm$0.13 & $+$0.60$\pm$0.07 & $+$1.20$\pm$0.21 & $-$0.15$\pm$0.11 & $+$0.16$\pm$0.08 & $+$0.07$\pm$0.04 & $+$0.23$\pm$0.04 \\
            \bottomrule
        \end{tabular}}
    } \\
    \subfloat[]{\label{app-subtab:acc_enp}
        \resizebox{0.95\columnwidth}{!}{
        \begin{tabular}{@{}lc|c|c|c|c|c|c|c|c|c@{}}
            \toprule
            & dut. & eng. & est. & fin. & fre. & ger. & lat. & pol. & swe. & all \\
            \midrule
            CharNet + ours & 78$\pm$5.2 & 80$\pm$3.2 & 81$\pm$2.0 & 77$\pm$4.7 & 73$\pm$1.8 & 71$\pm$3.9 & 76$\pm$5.0 & 70$\pm$1.5 & 58$\pm$2.2 & 74$\pm$1.2 \\
            \midrule
            Google + ours & 22$\pm$10.5 & 30$\pm$5.7 & 63$\pm$4.0 & 11$\pm$26.0 & 11$\pm$4.8 & 5$\pm$8.0 & 65$\pm$10.7 & 42$\pm$3.9 & 32$\pm$11.1 & 31$\pm$3.8 \\
            \bottomrule
        \end{tabular}}
    }
    \caption{\protect\subref{app-subtab:cer_enp} CER of the base OCR (resp.\ $\Delta$CER from the base to our correction, where negative values indicate improvement), in \%, computed on each sub-collection of ENP (as average over lines weighted by line length), then averaged (unweighted). Shown values indicate the 95\% confidence interval obtained by bootstrapping, sub-collection-wise, over $10^4$ resamples. \protect\subref{app-subtab:acc_enp} Correction accuracy, in \%, as defined in \Cref{eq:acc} (heuristic estimation).}
    \label{app-tab:metrics_enp}
\end{table}

\subsection{Ablation study: GMM clustering}

\begin{figure}
\centering
\begin{tabular}{cccc}
     & Berrutti high & Berrutti medium & Berrutti low \\
    \rotatebox[origin=c]{90}{CharNet}
        & \parbox[c]{0.3\columnwidth}{\includeinkscape[width=0.3\columnwidth]{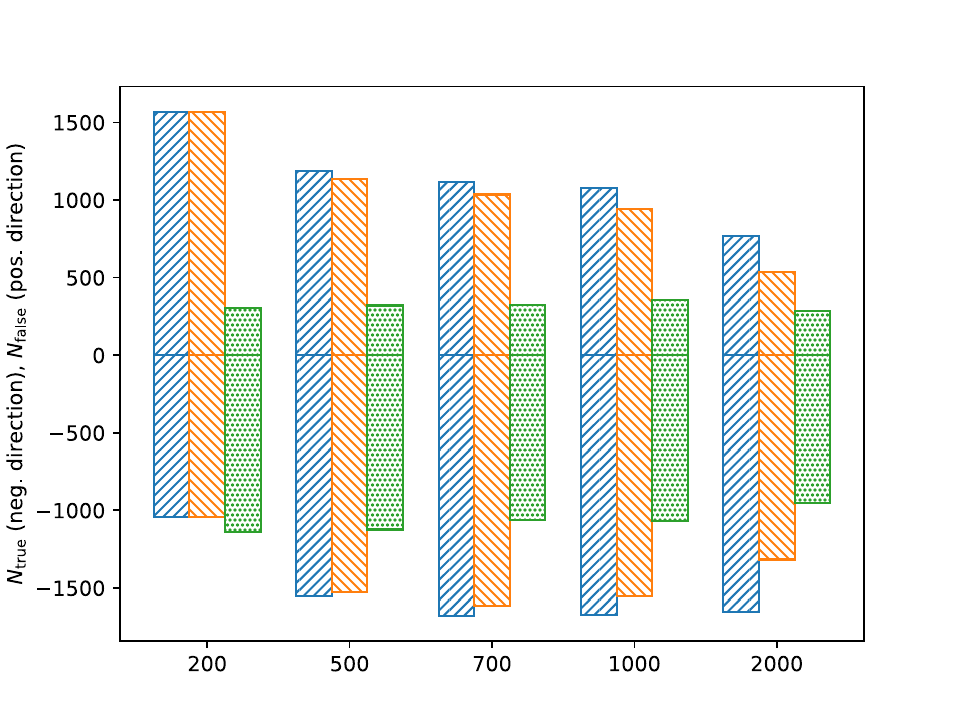}}
        & \parbox[c]{0.3\columnwidth}{\includeinkscape[width=0.3\columnwidth]{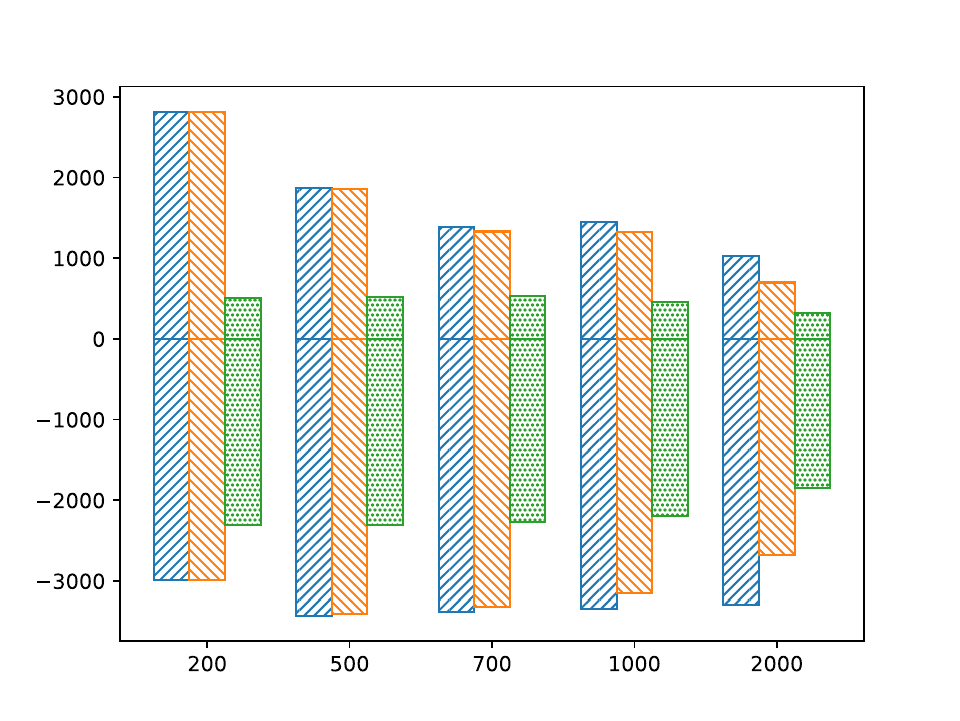}}
        & \parbox[c]{0.3\columnwidth}{\includeinkscape[width=0.3\columnwidth]{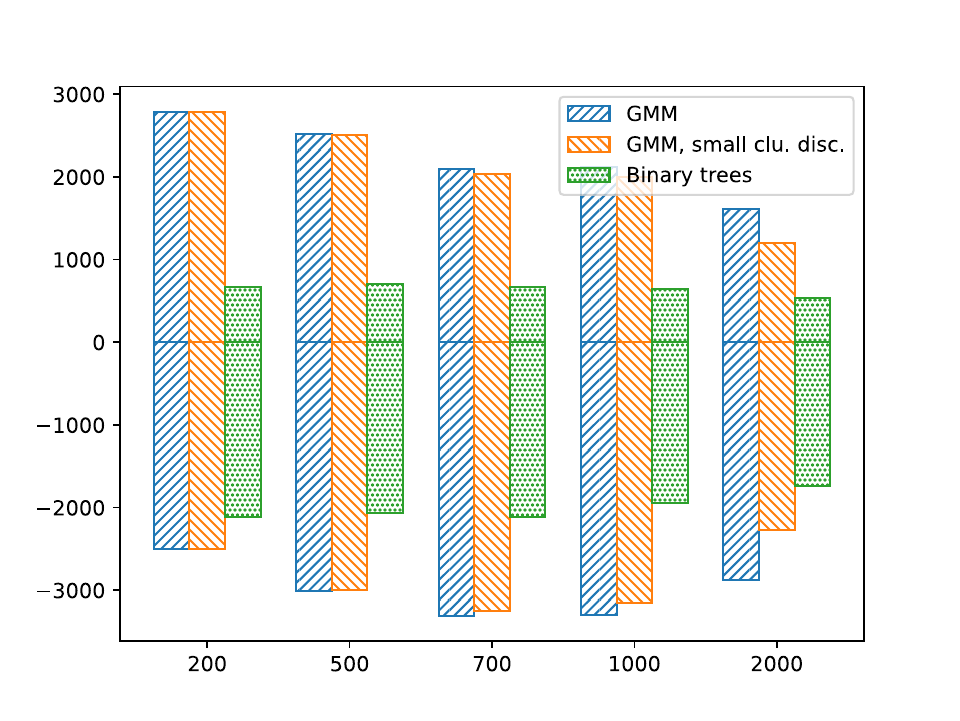}} \\
    \rotatebox[origin=c]{90}{Google}
        & \parbox[c]{0.3\columnwidth}{\includeinkscape[width=0.3\columnwidth]{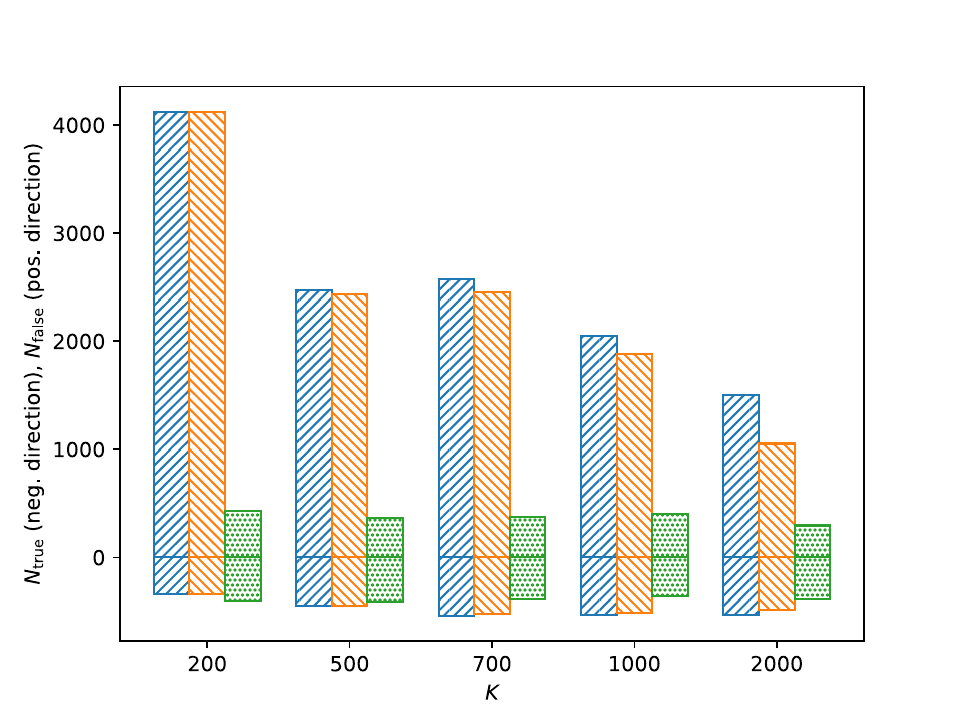}}
        & \parbox[c]{0.3\columnwidth}{\includeinkscape[width=0.3\columnwidth]{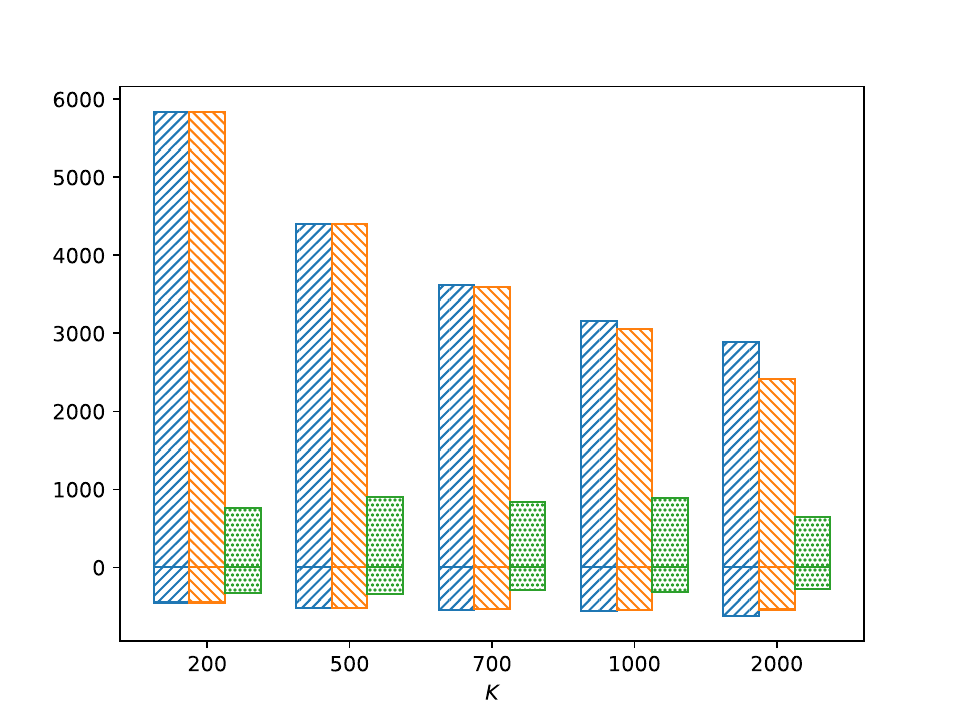}}
        & \parbox[c]{0.3\columnwidth}{\includeinkscape[width=0.3\columnwidth]{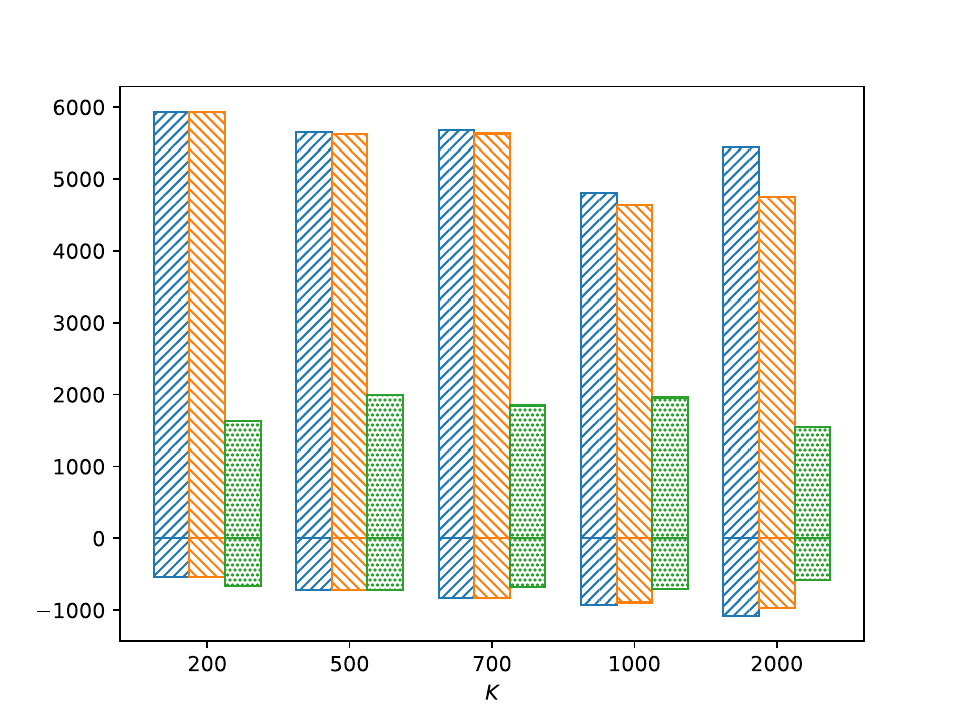}} \\
\end{tabular}
\caption{$N_\mathrm{false}$ (resp. $-N_\mathrm{true}$) on the positive (resp. negative) scale, heuristically estimated on each
sub-collection of Berrutti as in \Cref{eq:acc}. The algorithm is run for different values of $K$, in one of three ablated versions: using the clustering output of the GMM step, using the same output from which small clusters (fewer than $n_\mathrm{min}$ elements) are discarded, and using the output of the binary tree refinement (\ie unablated).}
\label{app-tab:ablation_gmm_berruti}
\end{figure}

\begin{figure}
\centering
\begin{tabular}{cccc}
     & ENP dutch & ENP english & ENP estonian \\
    \rotatebox[origin=c]{90}{CharNet}
        & \parbox[c]{0.3\columnwidth}{\includeinkscape[width=0.3\columnwidth]{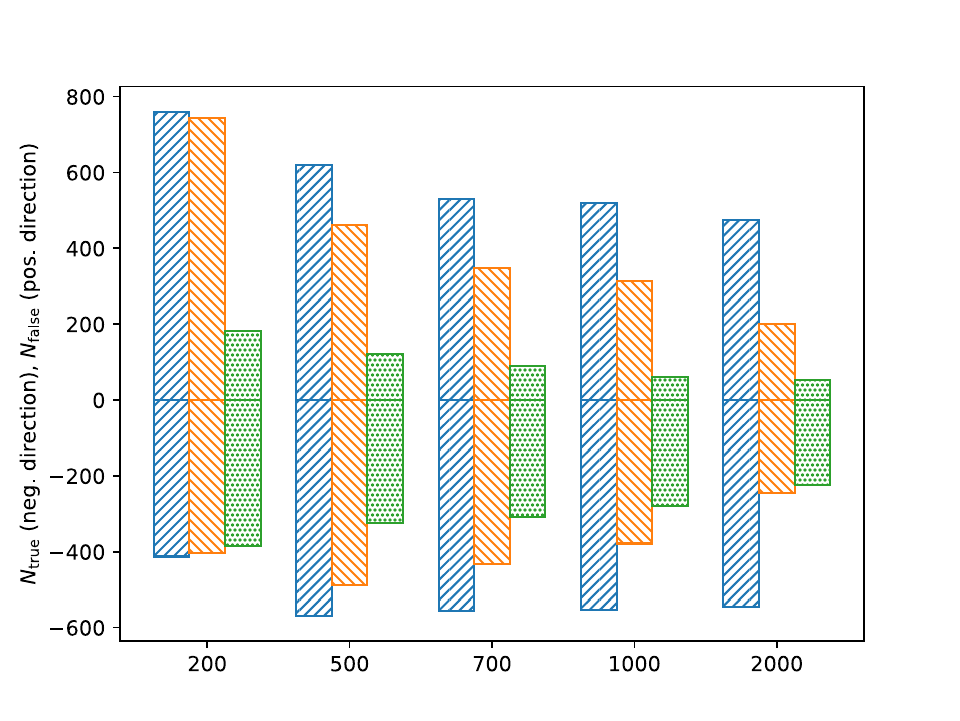}}
        & \parbox[c]{0.3\columnwidth}{\includeinkscape[width=0.3\columnwidth]{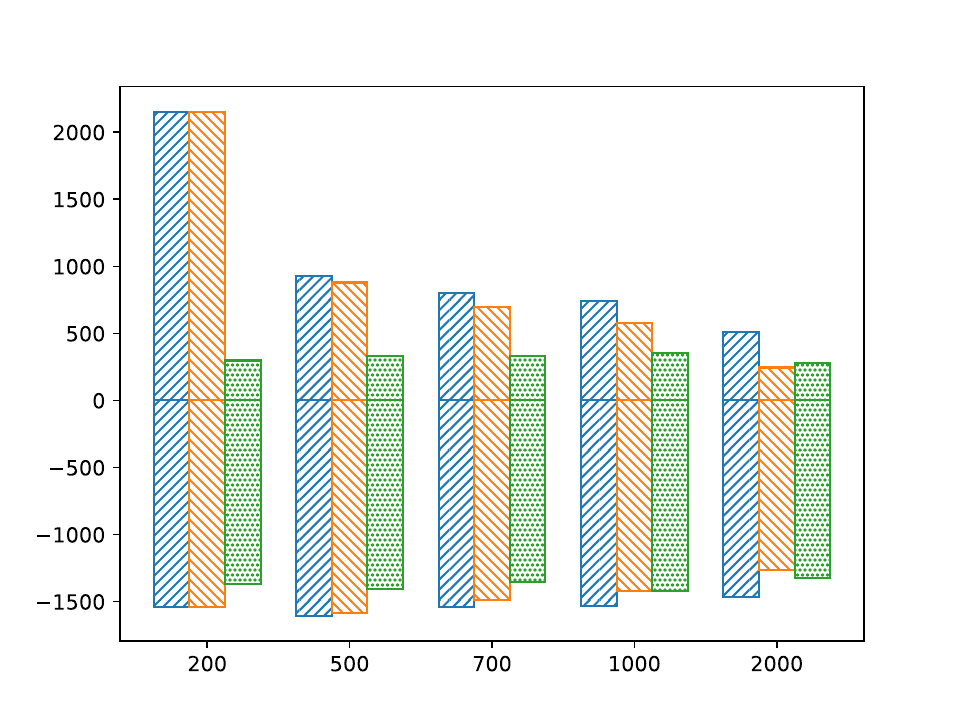}}
        & \parbox[c]{0.3\columnwidth}{\includeinkscape[width=0.3\columnwidth]{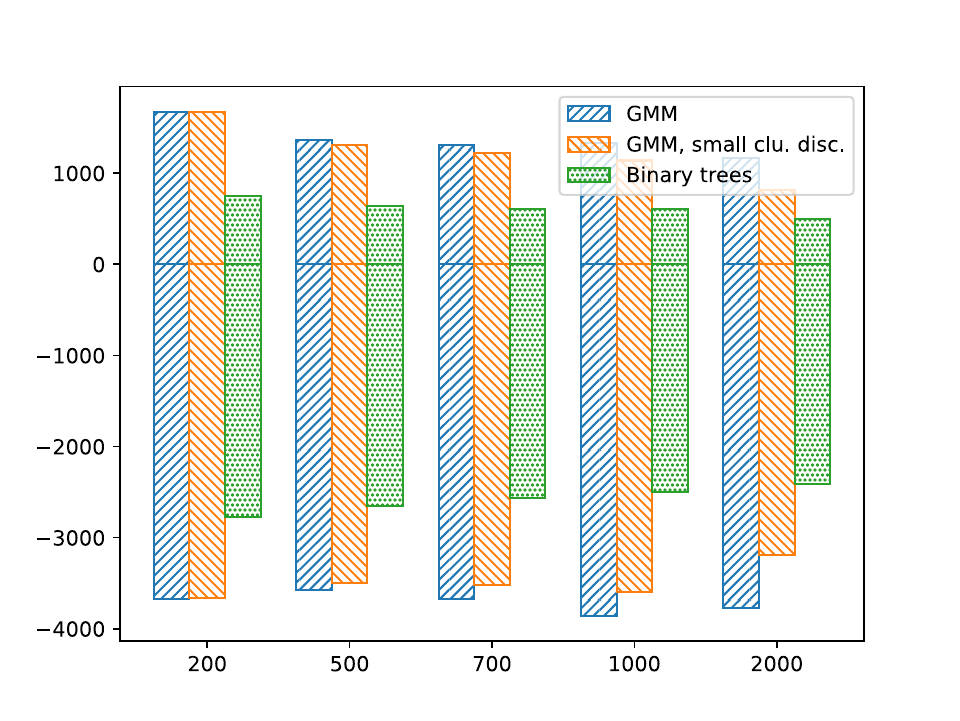}} \\
    \rotatebox[origin=c]{90}{Google}
        & \parbox[c]{0.3\columnwidth}{\includeinkscape[width=0.3\columnwidth]{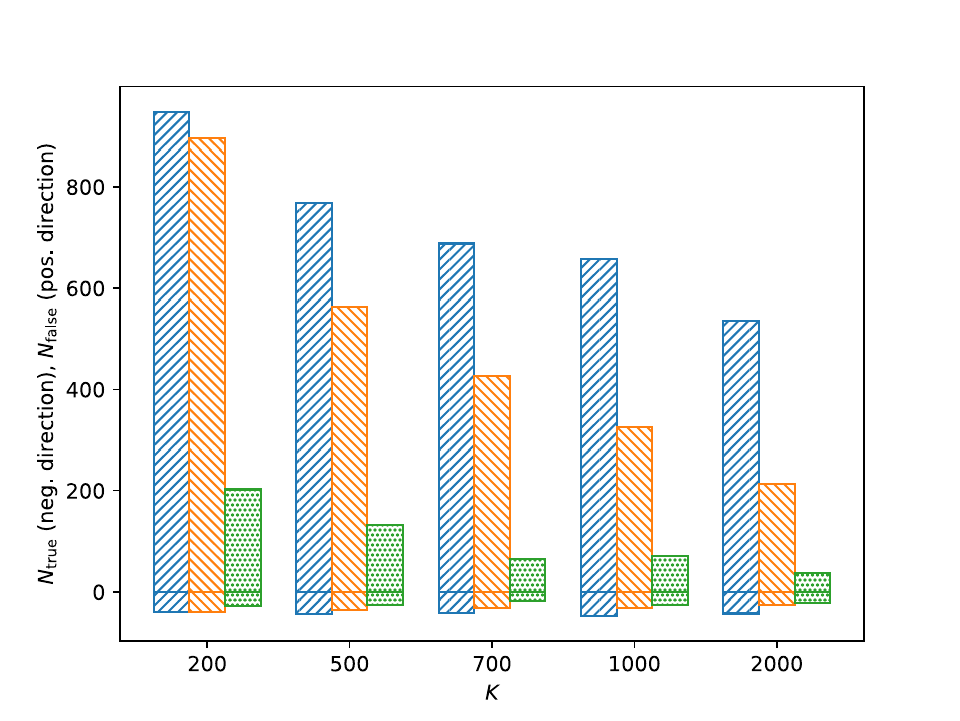}}
        & \parbox[c]{0.3\columnwidth}{\includeinkscape[width=0.3\columnwidth]{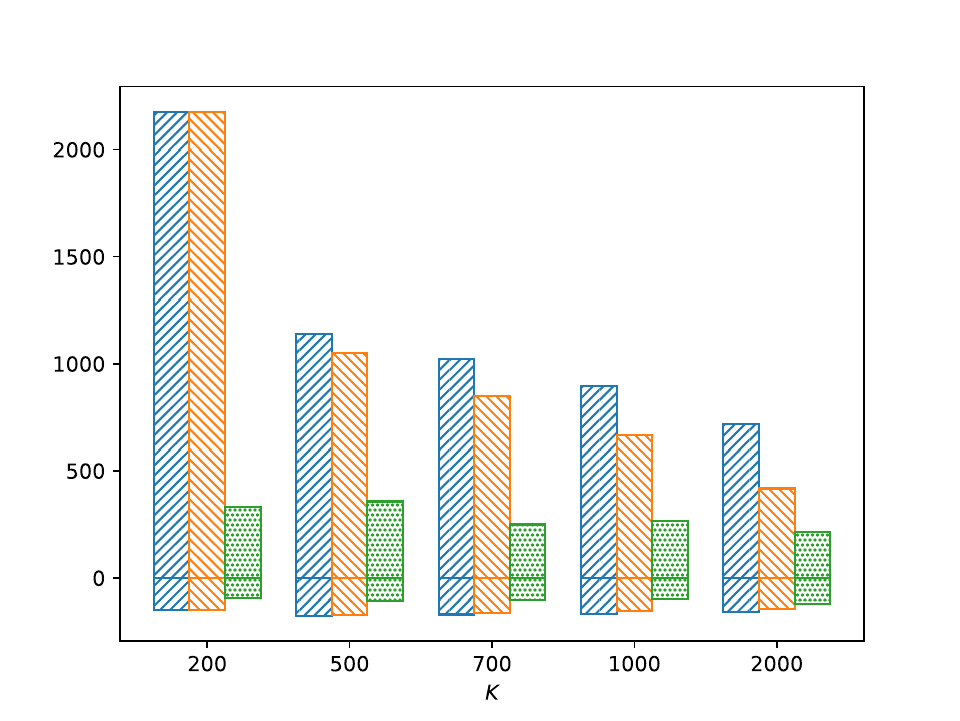}}
        & \parbox[c]{0.3\columnwidth}{\includeinkscape[width=0.3\columnwidth]{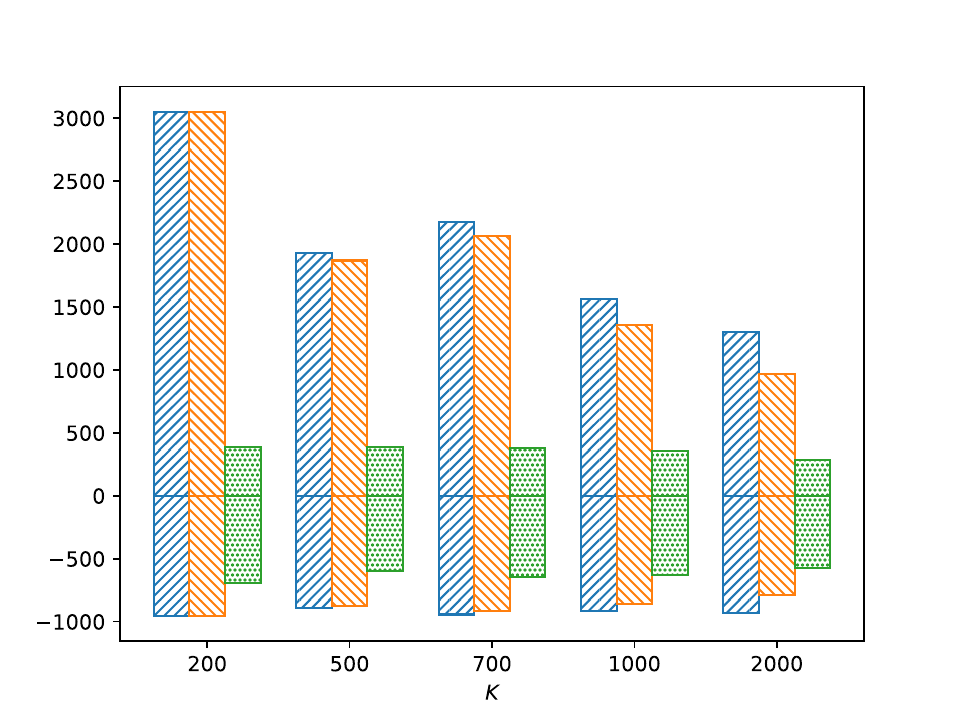}} \\
     & ENP finnish & ENP french  & ENP german \\
    \rotatebox[origin=c]{90}{CharNet}
        & \parbox[c]{0.3\columnwidth}{\includeinkscape[width=0.3\columnwidth]{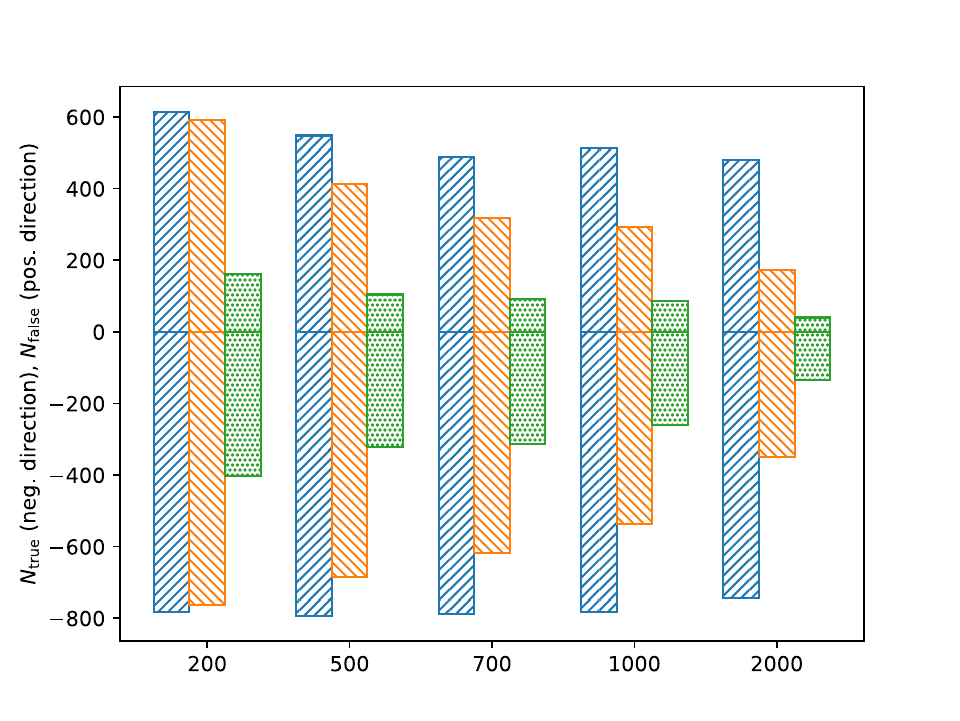}}
        & \parbox[c]{0.3\columnwidth}{\includeinkscape[width=0.3\columnwidth]{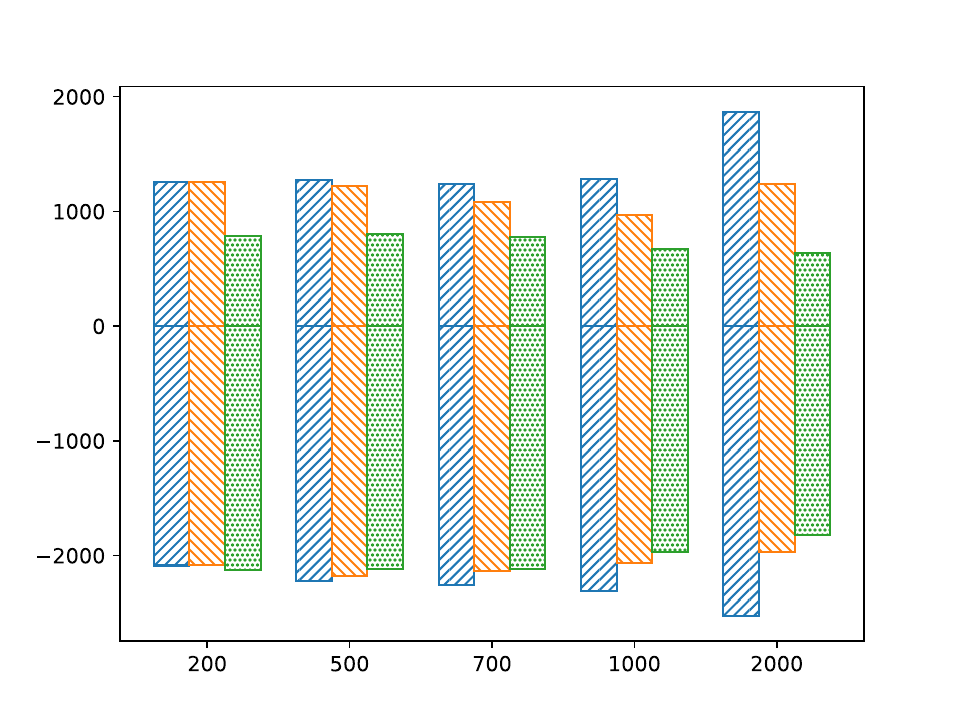}}
        & \parbox[c]{0.3\columnwidth}{\includeinkscape[width=0.3\columnwidth]{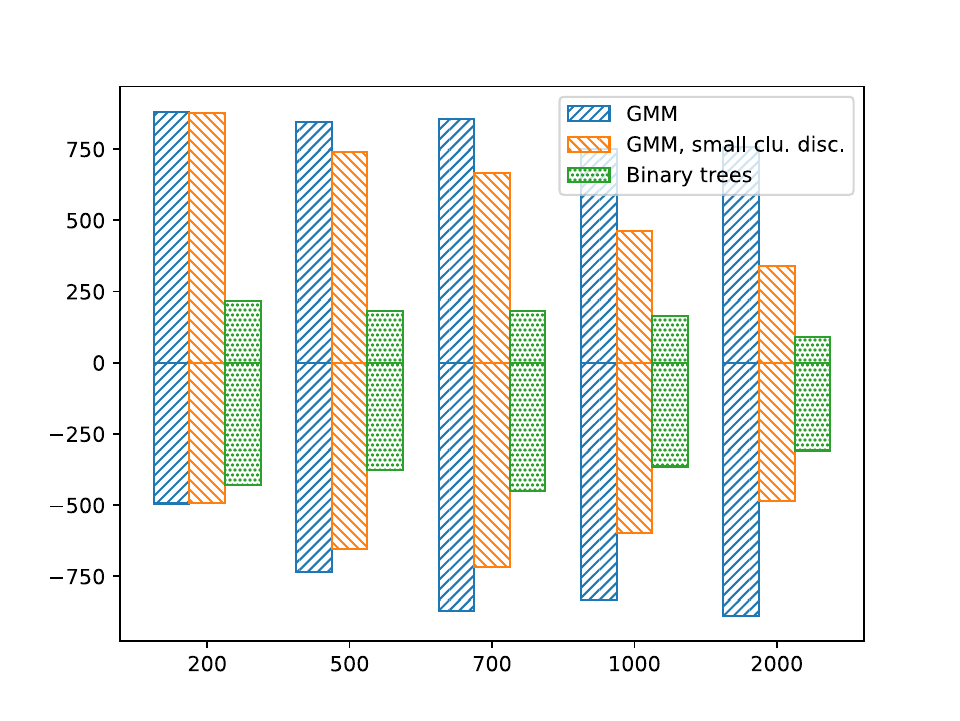}} \\
    \rotatebox[origin=c]{90}{Google}
        & \parbox[c]{0.3\columnwidth}{\includeinkscape[width=0.3\columnwidth]{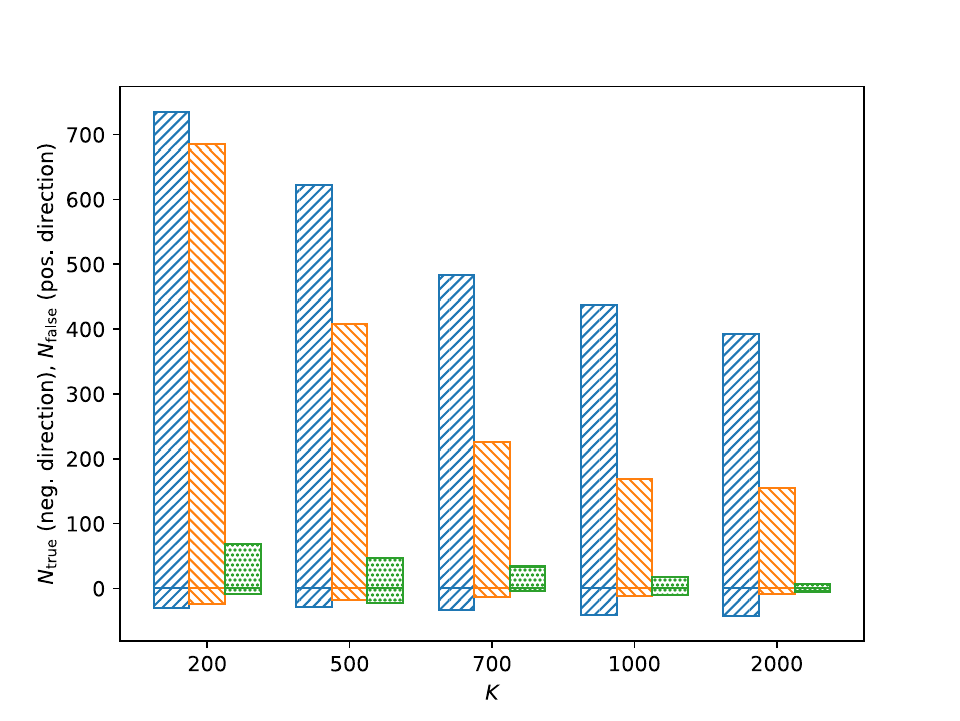}}
        & \parbox[c]{0.3\columnwidth}{\includeinkscape[width=0.3\columnwidth]{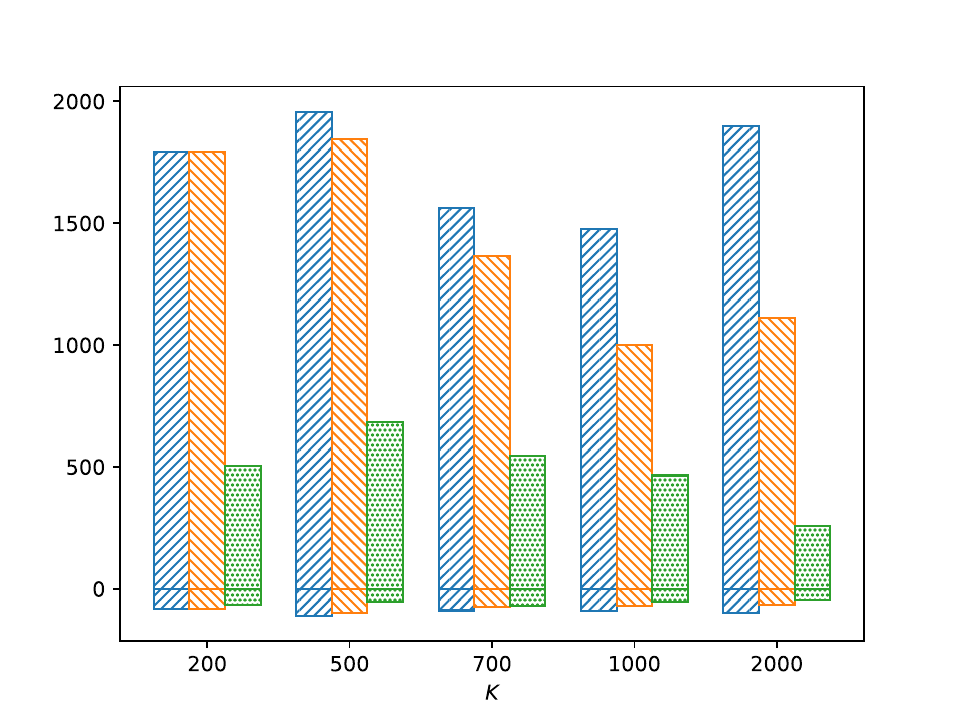}}
        & \parbox[c]{0.3\columnwidth}{\includeinkscape[width=0.3\columnwidth]{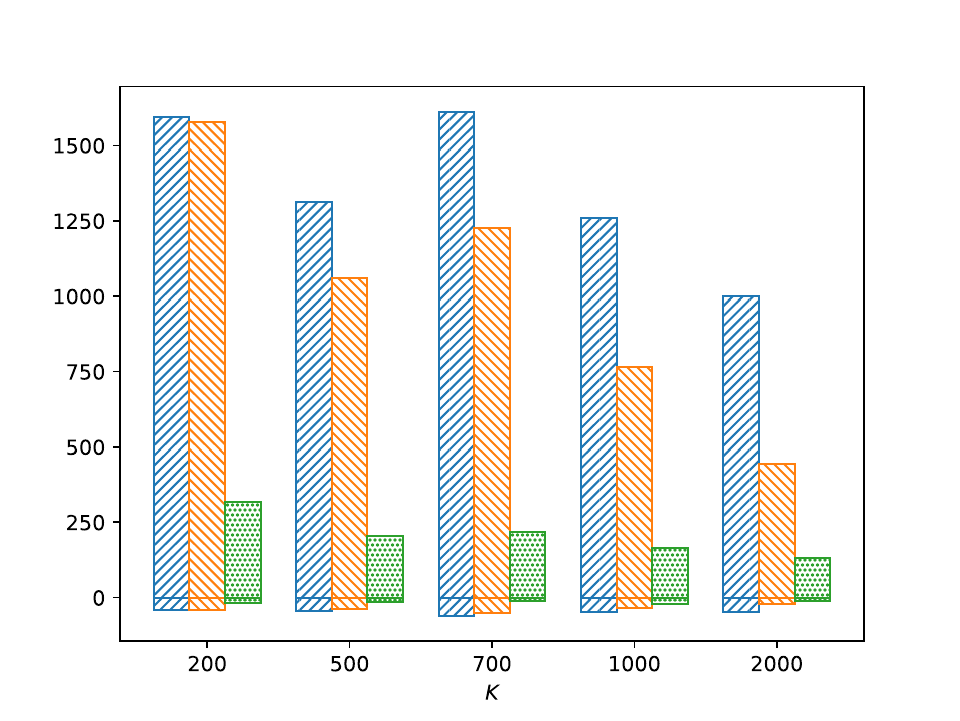}} \\
     & ENP latvian & ENP polish  & ENP swedish \\
    \rotatebox[origin=c]{90}{CharNet}
        & \parbox[c]{0.3\columnwidth}{\includeinkscape[width=0.3\columnwidth]{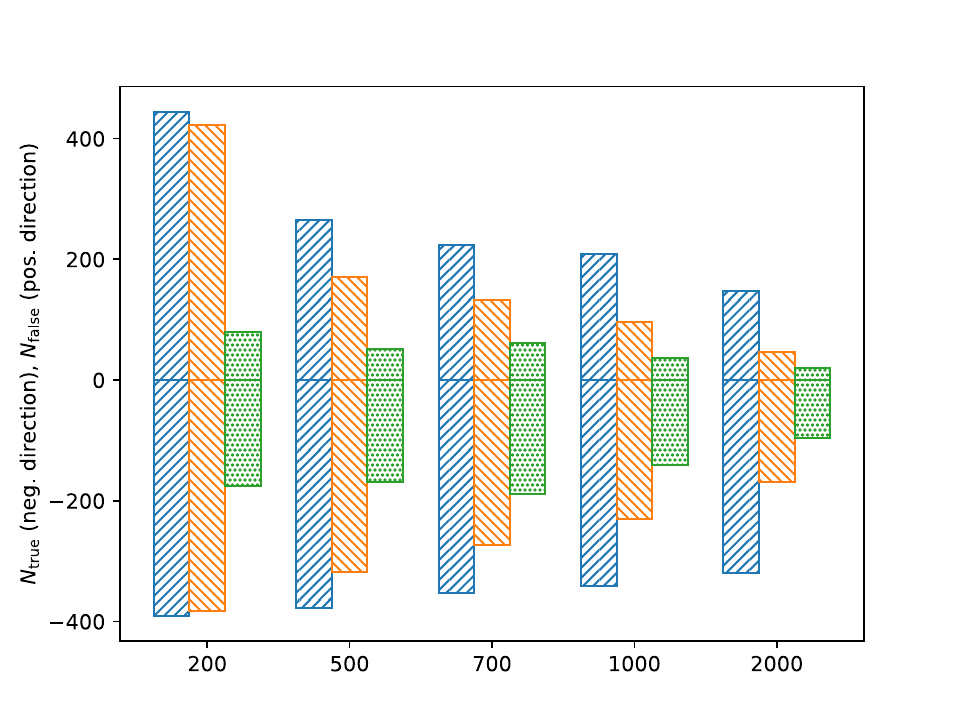}}
        & \parbox[c]{0.3\columnwidth}{\includeinkscape[width=0.3\columnwidth]{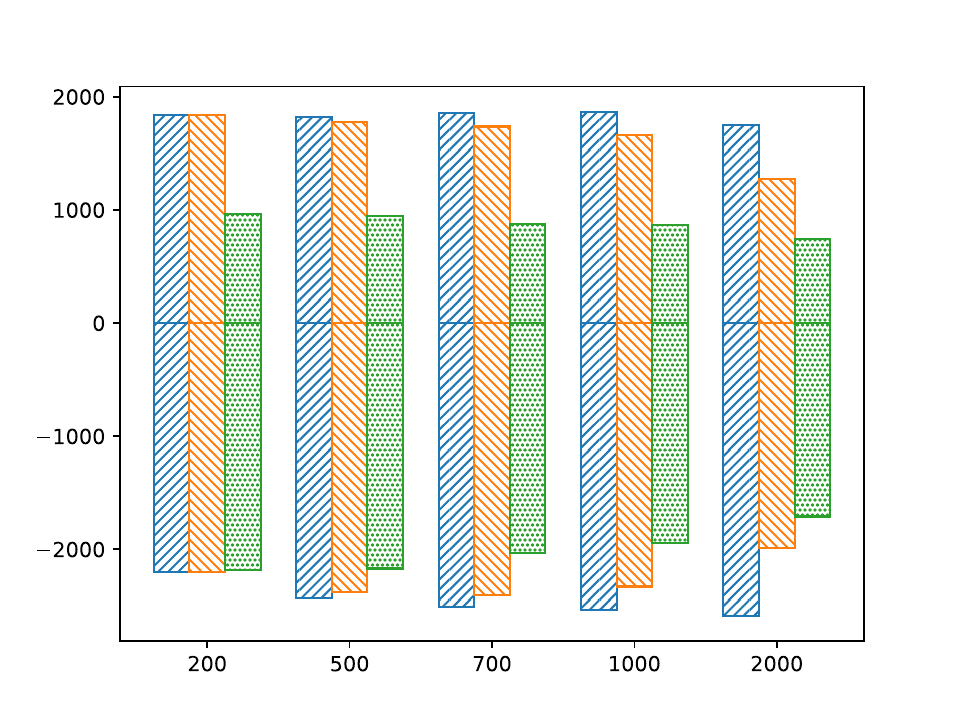}}
        & \parbox[c]{0.3\columnwidth}{\includeinkscape[width=0.3\columnwidth]{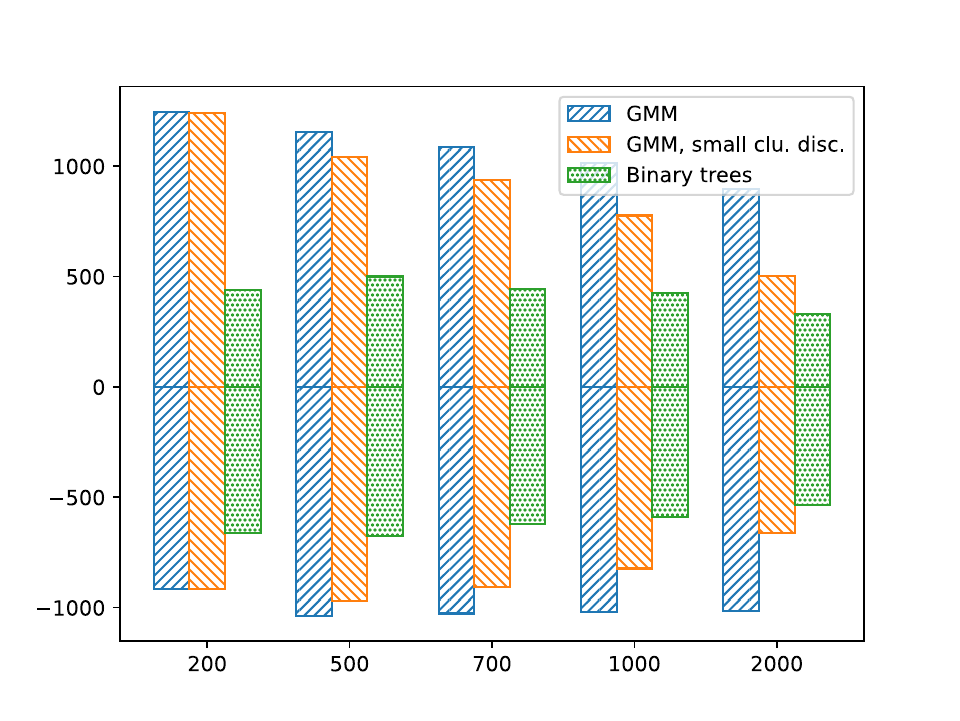}} \\
    \rotatebox[origin=c]{90}{Google}
        & \parbox[c]{0.3\columnwidth}{\includeinkscape[width=0.3\columnwidth]{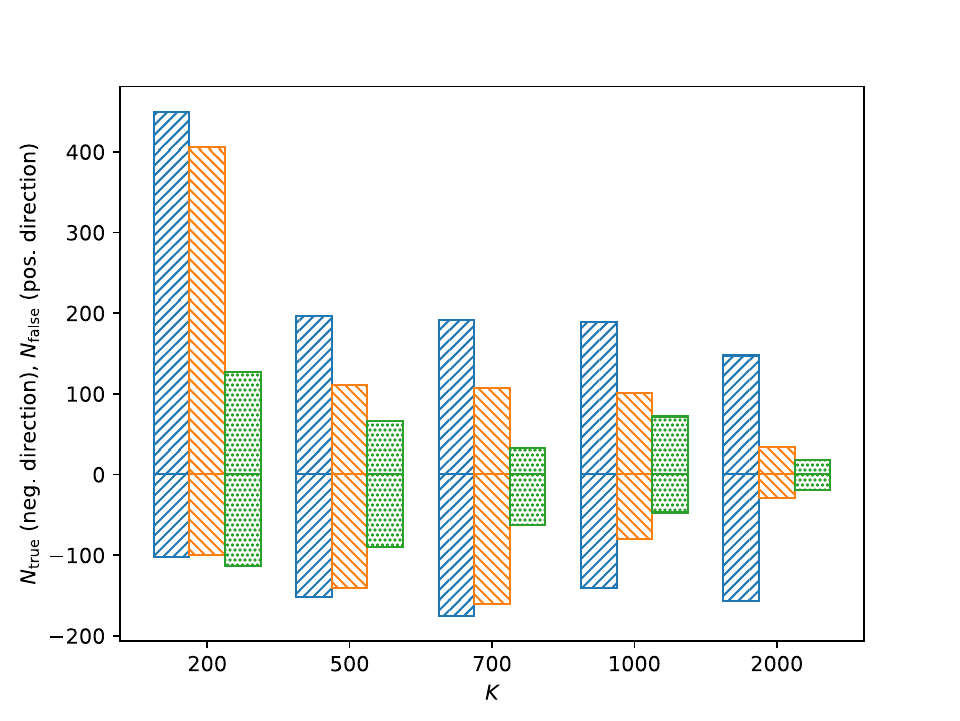}}
        & \parbox[c]{0.3\columnwidth}{\includeinkscape[width=0.3\columnwidth]{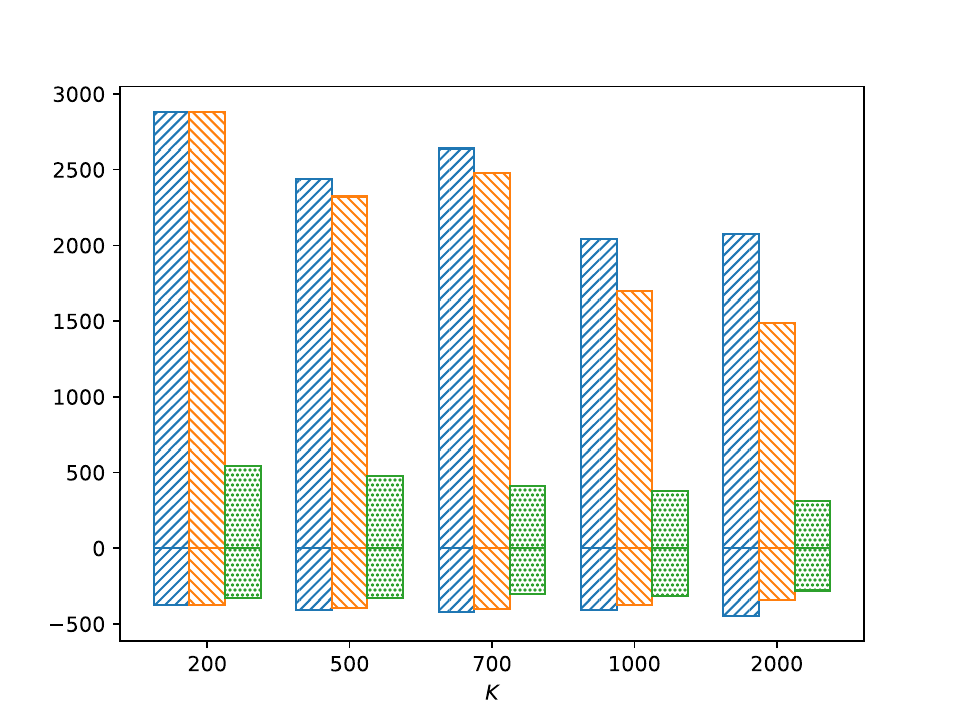}}
        & \parbox[c]{0.3\columnwidth}{\includeinkscape[width=0.3\columnwidth]{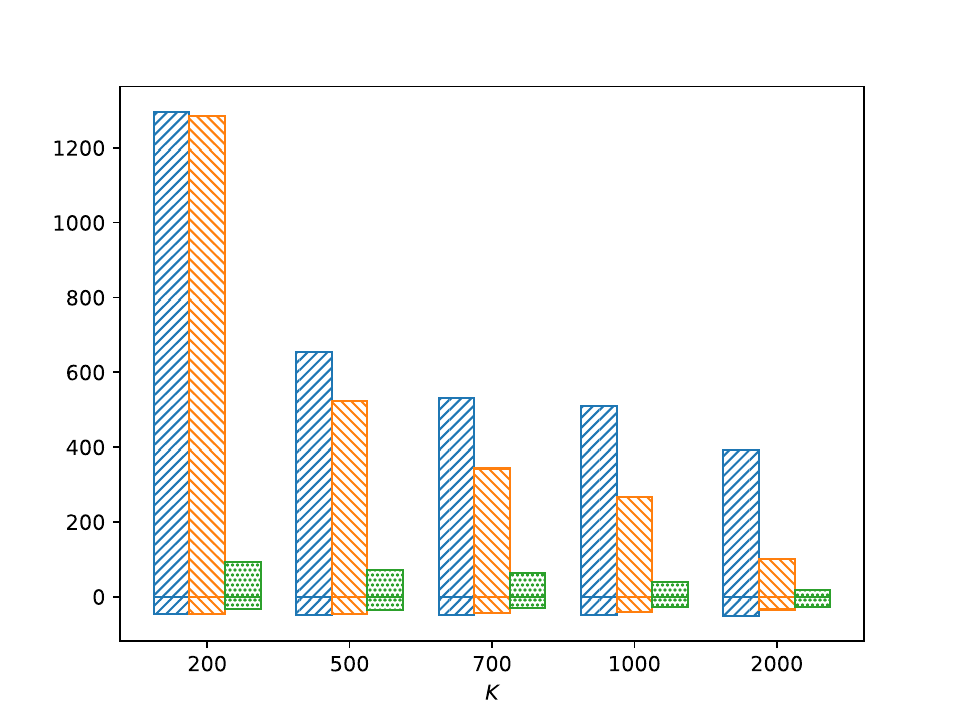}} \\
\end{tabular}
\caption{$N_\mathrm{false}$ (resp. $-N_\mathrm{true}$) on the positive (resp. negative) scale, heuristically estimated on each
sub-collection of ENP as in \Cref{eq:acc}. The algorithm is run for different values of $K$, in one of three ablated versions: using the clustering output of the GMM step, using the same output from which small clusters (fewer than $n_\mathrm{min}$ elements) are discarded, and using the output of the binary tree refinement (\ie unablated).}
\label{app-tab:ablation_gmm_enp}
\end{figure}

We provide here the detailed experimental data of the ablation study, broken down by sub-collection (and not aggregated by dataset). \Cref{app-tab:ablation_gmm_berruti,app-tab:ablation_gmm_enp} correspond to \Cref{fig:ablation_gmm}, and \Cref{app-tab:ablation_gmm_numclusters} corresponds to \Cref{tab:ablation_gmm_numclusters}. They generally confirm the observations of the main paper about performance and robustness, except for a few situations (base OCR Google and sub-collections dutch, latvian, finnish, swedish of ENP) where robustness breaks down before $K=2000$.

\begin{table*}
\centering
\subfloat[]{ \label{app-subtab:numclu}
    \resizebox{0.48\columnwidth}{!}{
    \begin{tabular}{@{}ll|ccccc|ccccc|ccccc@{}}
        \toprule
        & Dataset, collection & \multicolumn{5}{|c}{Berrutti high} & \multicolumn{5}{|c}{Berrutti medium} & \multicolumn{5}{|c}{Berrutti low}\\
        & $K$ & 200 & 500 & 700 & 1000 & 2000 & 200 & 500 & 700 & 1000 & 2000 & 200 & 500 & 700 & 1000 & 2000\\
        \midrule
        & GMM & 199 & 499 & 696 & 986 & 1924 & 199 & 499 & 698 & 996 & 1962 & 199 & 499 & 697 & 995 & 1936 \\
        & GMM, small clu. disc. & 199 & 431 & 518 & 594 & 803 & 197 & 475 & 601 & 728 & 854 & 198 & 475 & 586 & 701 & 761 \\
        & Binary trees & 1685 & 1737 & 1747 & 1709 & 1696 & 1235 & 1178 & 1237 & 1205 & 1152 & 909 & 946 & 954 & 953 & 852 \\
        \midrule
        & GMM & 199 & 499 & 697 & 995 & 1936 & 199 & 499 & 699 & 995 & 1976 & 199 & 499 & 699 & 995 & 1967 \\
        & GMM, small clu. disc. & 199 & 451 & 559 & 690 & 773 & 199 & 497 & 670 & 864 & 1170 & 199 & 486 & 644 & 844 & 1122 \\
        & Binary trees & 1871 & 1881 & 1855 & 1868 & 1796 & 1460 & 1530 & 1543 & 1558 & 1487 & 1381 & 1406 & 1415 & 1408 & 1399 \\
        \bottomrule
        \toprule
        & Dataset, collection & \multicolumn{5}{|c}{ENP dutch} & \multicolumn{5}{|c}{ENP english} & \multicolumn{5}{|c}{ENP estonian}\\
        & $K$ & 200 & 500 & 700 & 1000 & 2000 & 200 & 500 & 700 & 1000 & 2000 & 200 & 500 & 700 & 1000 & 2000\\
        \midrule
        & GMM & 197 & 490 & 685 & 958 & 1751 & 199 & 492 & 675 & 946 & 1757 & 199 & 499 & 698 & 990 & 1888 \\
        & GMM, small clu. disc. & 176 & 198 & 201 & 180 & 148 & 195 & 360 & 387 & 391 & 396 & 197 & 404 & 463 & 507 & 544 \\
        & Binary trees & 292 & 259 & 269 & 237 & 191 & 1211 & 1173 & 1211 & 1234 & 1162 & 936 & 977 & 921 & 943 & 871 \\
        \midrule
        & GMM & 198 & 485 & 675 & 951 & 1766 & 199 & 487 & 671 & 935 & 1726 & 199 & 498 & 695 & 983 & 1901 \\
        & GMM, small clu. disc. & 165 & 195 & 193 & 182 & 146 & 195 & 307 & 326 & 326 & 325 & 198 & 436 & 517 & 579 & 620 \\
        & Binary trees & 294 & 261 & 258 & 226 & 165 & 912 & 882 & 938 & 907 & 824 & 1075 & 1104 & 1093 & 1061 & 1008 \\
        \bottomrule
        \toprule
        & Dataset, collection & \multicolumn{5}{|c}{ENP finnish} & \multicolumn{5}{|c}{ENP french} & \multicolumn{5}{|c}{ENP german}\\
        & $K$ & 200 & 500 & 700 & 1000 & 2000 & 200 & 500 & 700 & 1000 & 2000 & 200 & 500 & 700 & 1000 & 2000\\
        \midrule
        & GMM & 198 & 482 & 650 & 902 & 1568 & 199 & 498 & 698 & 992 & 1941 & 199 & 490 & 677 & 936 & 1673 \\
        & GMM, small clu. disc. & 146 & 165 & 142 & 134 & 97 & 198 & 398 & 436 & 424 & 399 & 179 & 222 & 205 & 191 & 124 \\
        & Binary trees & 209 & 201 & 176 & 170 & 123 & 735 & 752 & 700 & 681 & 619 & 197 & 172 & 177 & 162 & 132 \\
        \midrule
        & GMM & 199 & 486 & 659 & 899 & 1534 & 199 & 497 & 695 & 987 & 1943 & 198 & 492 & 673 & 925 & 1618 \\
        & GMM, small clu. disc. & 141 & 137 & 129 & 125 & 63 & 198 & 385 & 436 & 462 & 413 & 175 & 220 & 209 & 180 & 97 \\
        & Binary trees & 168 & 179 & 168 & 145 & 91 & 681 & 726 & 712 & 722 & 612 & 192 & 181 & 178 & 162 & 115 \\
        \bottomrule
        \toprule
        & Dataset, collection & \multicolumn{5}{|c}{ENP latvian} & \multicolumn{5}{|c}{ENP polish} & \multicolumn{5}{|c}{ENP swedish}\\
        & $K$ & 200 & 500 & 700 & 1000 & 2000 & 200 & 500 & 700 & 1000 & 2000 & 200 & 500 & 700 & 1000 & 2000\\
        \midrule
        & GMM & 198 & 479 & 650 & 888 & 1637 & 199 & 496 & 696 & 990 & 1920 & 199 & 498 & 696 & 970 & 1871 \\
        & GMM, small clu. disc. & 148 & 181 & 213 & 193 & 193 & 198 & 406 & 461 & 518 & 549 & 192 & 317 & 333 & 345 & 293 \\
        & Binary trees & 229 & 245 & 224 & 220 & 174 & 890 & 906 & 882 & 880 & 820 & 531 & 560 & 519 & 494 & 439 \\
        \midrule
        & GMM & 198 & 477 & 652 & 897 & 1661 & 199 & 499 & 696 & 989 & 1928 & 199 & 494 & 685 & 975 & 1842 \\
        & GMM, small clu. disc. & 146 & 187 & 184 & 208 & 173 & 197 & 381 & 464 & 471 & 459 & 188 & 296 & 286 & 296 & 268 \\
        & Binary trees & 249 & 240 & 208 & 204 & 151 & 784 & 757 & 755 & 752 & 675 & 486 & 496 & 500 & 497 & 431 \\
        \bottomrule
    \end{tabular}}}
\subfloat[]{ \label{app-subtab:propchars}
    \resizebox{0.48\columnwidth}{!}{
    \begin{tabular}{@{}ll|ccccc|ccccc|ccccc@{}}
        \toprule
        & Dataset, collection & \multicolumn{5}{|c}{Berrutti high} & \multicolumn{5}{|c}{Berrutti medium} & \multicolumn{5}{|c}{Berrutti low}\\
        & $K$ & 200 & 500 & 700 & 1000 & 2000 & 200 & 500 & 700 & 1000 & 2000 & 200 & 500 & 700 & 1000 & 2000\\
        \midrule
        & GMM, small clu. disc. & 100 & 99 & 99 & 97 & 93 & 100 & 100 & 99 & 96 & 87 & 100 & 100 & 98 & 95 & 83 \\
        & Binary trees & 79 & 78 & 78 & 78 & 76 & 77 & 76 & 76 & 75 & 68 & 67 & 69 & 68 & 67 & 60 \\
        \midrule
        & GMM, small clu. disc. & 100 & 100 & 99 & 98 & 93 & 100 & 100 & 100 & 99 & 92 & 100 & 100 & 99 & 98 & 91 \\
        & Binary trees & 78 & 78 & 79 & 77 & 75 & 75 & 75 & 75 & 75 & 69 & 69 & 70 & 70 & 70 & 66 \\
        \bottomrule
        \toprule
        & Dataset, collection & \multicolumn{5}{|c}{ENP dutch} & \multicolumn{5}{|c}{ENP english} & \multicolumn{5}{|c}{ENP estonian}\\
        & $K$ & 200 & 500 & 700 & 1000 & 2000 & 200 & 500 & 700 & 1000 & 2000 & 200 & 500 & 700 & 1000 & 2000\\
        \midrule
        & GMM, small clu. disc. & 99 & 88 & 84 & 77 & 61 & 100 & 99 & 98 & 96 & 93 & 100 & 98 & 97 & 93 & 85 \\
        & Binary trees & 76 & 72 & 67 & 64 & 52 & 82 & 82 & 81 & 81 & 79 & 77 & 75 & 74 & 72 & 68 \\
        \midrule
        & GMM, small clu. disc. & 98 & 88 & 83 & 76 & 57 & 100 & 98 & 97 & 95 & 92 & 100 & 99 & 98 & 95 & 87 \\
        & Binary trees & 78 & 73 & 68 & 64 & 50 & 88 & 88 & 87 & 86 & 84 & 78 & 77 & 75 & 75 & 70 \\
        \bottomrule
        \toprule
        & Dataset, collection & \multicolumn{5}{|c}{ENP finnish} & \multicolumn{5}{|c}{ENP french} & \multicolumn{5}{|c}{ENP german}\\
        & $K$ & 200 & 500 & 700 & 1000 & 2000 & 200 & 500 & 700 & 1000 & 2000 & 200 & 500 & 700 & 1000 & 2000\\
        \midrule
        & GMM, small clu. disc. & 96 & 83 & 76 & 68 & 50 & 100 & 98 & 96 & 93 & 83 & 99 & 85 & 77 & 70 & 51 \\
        & Binary trees & 54 & 52 & 48 & 42 & 35 & 84 & 82 & 82 & 79 & 73 & 47 & 43 & 45 & 37 & 33 \\
        \midrule
        & GMM, small clu. disc. & 96 & 79 & 72 & 65 & 43 & 100 & 98 & 96 & 93 & 82 & 98 & 84 & 75 & 65 & 45 \\
        & Binary trees & 53 & 54 & 51 & 43 & 33 & 87 & 85 & 83 & 80 & 73 & 50 & 47 & 46 & 42 & 32 \\
        \bottomrule
        \toprule
        & Dataset, collection & \multicolumn{5}{|c}{ENP latvian} & \multicolumn{5}{|c}{ENP polish} & \multicolumn{5}{|c}{ENP swedish}\\
        & $K$ & 200 & 500 & 700 & 1000 & 2000 & 200 & 500 & 700 & 1000 & 2000 & 200 & 500 & 700 & 1000 & 2000\\
        \midrule
        & GMM, small clu. disc. & 97 & 88 & 83 & 74 & 47 & 100 & 99 & 96 & 94 & 84 & 100 & 96 & 93 & 89 & 77 \\
        & Binary trees & 69 & 60 & 63 & 55 & 35 & 78 & 77 & 76 & 73 & 67 & 72 & 74 & 70 & 68 & 62 \\
        \midrule
        & GMM, small clu. disc. & 97 & 86 & 79 & 70 & 39 & 100 & 98 & 96 & 92 & 80 & 100 & 96 & 92 & 88 & 76 \\
        & Binary trees & 74 & 63 & 60 & 54 & 31 & 83 & 80 & 80 & 78 & 69 & 79 & 77 & 76 & 74 & 66 \\
        \bottomrule
    \end{tabular}}}
\caption{Complementary data for the ablation study (see \cref{app-tab:ablation_gmm_berruti,app-tab:ablation_gmm_enp}). \protect\subref{app-subtab:numclu} Number of clusters. \protect\subref{app-subtab:propchars} Proportion, in \%, of characters in retained (non-discarded) clusters.}
\label{app-tab:ablation_gmm_numclusters}
\end{table*}

\end{document}